\documentclass[letterpaper]{article} 
\usepackage[preprint]{aaai2027}  
\usepackage[hyphens]{url}  
\usepackage{graphicx} 
\def\UrlFont{\rm}  
\usepackage{natbib}  
\usepackage{caption} 
\usepackage{algorithm}
\usepackage{algorithmic}

\usepackage{newfloat}
\usepackage{listings}
\DeclareCaptionStyle{ruled}{labelfont=normalfont,labelsep=colon,strut=off} 
\floatstyle{ruled}
\newfloat{listing}{tb}{lst}{}
\floatname{listing}{Listing}

\usepackage{booktabs}
\usepackage{amsmath}
\usepackage{amssymb}
\usepackage{multirow}
\usepackage[most]{tcolorbox}

\definecolor{promptbg}{HTML}{EEF4FB}
\definecolor{promptframe}{HTML}{5E81AC}
\definecolor{casebg}{HTML}{F5F8FC}
\definecolor{caseframe}{HTML}{7A9BC2}
\newtcblisting{promptbox}[1][]{
  enhanced,
  breakable,
  listing only,
  colback=promptbg,
  colframe=promptframe,
  fonttitle=\bfseries\small,
  title={#1},
  boxrule=0.5pt,
  arc=2pt,
  left=6pt,
  right=6pt,
  top=4pt,
  bottom=4pt,
  listing options={
    basicstyle=\footnotesize\ttfamily,
    numbers=none,
    xleftmargin=0pt,
    aboveskip=0pt,
    belowskip=0pt,
    breaklines=true,
    columns=fullflexible,
    keepspaces=true,
    showstringspaces=false,
    tabsize=2
  }
}
\newtcolorbox{casebox}[1][]{
  enhanced,
  breakable,
  colback=casebg,
  colframe=caseframe,
  fonttitle=\bfseries\small,
  fontupper=\small,
  title={#1},
  boxrule=0.5pt,
  arc=2pt,
  left=6pt,
  right=6pt,
  top=4pt,
  bottom=4pt
}
\newcommand{\todo}[1]{\textbf{[TODO: #1]}}

\newcommand{\resourceLink}[2]{%
    \leavevmode
    \pdfstartlink attr {/Border [0 0 0]} user {%
        /Subtype /Link /A << /S /URI /URI (#1) >>%
    }#2\pdfendlink%
}

\title{SkillReason: Reasoning-Enhanced Agent Skill Retrieval for Implicit User Requests}
\author{
Donghong Jiang\textsuperscript{\rm 1},
Endian Lin\textsuperscript{\rm 1},
Luoping Cui\textsuperscript{\rm 1},
Hanqing Liu\textsuperscript{\rm 1},
Mingjie Liu\textsuperscript{\rm 1},\\
Fan Yang\textsuperscript{\rm 2,\rm 4},
Hong Wang\textsuperscript{\rm 1},
Zhao Yang\textsuperscript{\rm 3},
Chuang Zhu\textsuperscript{\rm 1}\textsuperscript{*}
}
\affiliations{
\textsuperscript{\rm 1}Beijing University of Posts and Telecommunications\\
\textsuperscript{\rm 2}Peking University\\
\textsuperscript{\rm 3}Beijing E-Hualu Information Technology Co., Ltd.\\
\textsuperscript{\rm 4}Beijing ZOYEN Technology Co., Ltd.\\
\{donghongjiang, ledgogo, lpcui, hanqingliu, LMJ,wanghong723,czhu\}@bupt.edu.cn,fanyang@zoyen.com.cn,
zhaoy01@ehualu.com\\
\mbox{%
\resourceLink{https://github.com/donghong1/SkillReason}{\raisebox{-0.12em}{\includegraphics[height=0.95em,trim=230 230 230 230,clip]{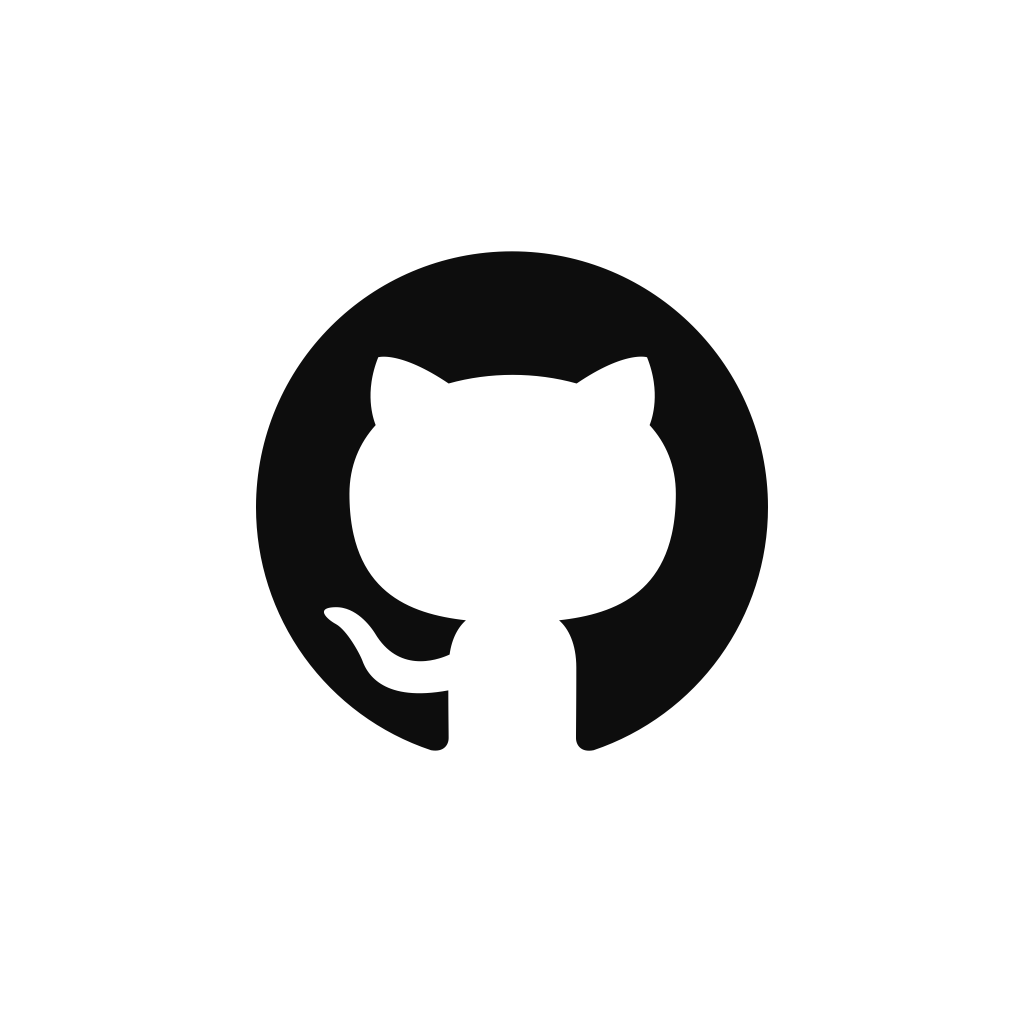}}\,\textbf{Code}}%
\qquad
\resourceLink{https://huggingface.co/donghongjiang}{\raisebox{-0.12em}{\includegraphics[height=0.95em]{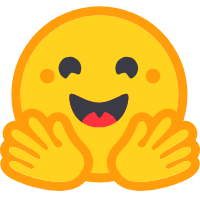}}\,\textbf{Models}}%
\qquad
\resourceLink{https://huggingface.co/datasets/donghongjiang/skillreason-bench}{\raisebox{-0.12em}{\includegraphics[height=0.95em]{Figures/huggingface.png}}\,\textbf{Benchmark}}%
}
}

\begin{document}

\maketitle

\begin{abstract}
Large language model agents increasingly rely on reusable skills to extend their capabilities beyond parametric knowledge. However, retrieving the appropriate skill from a large-scale library remains challenging because realistic user requests are often concise and underspecified, stating only the task goal while leaving the required capabilities and execution steps implicit. Existing benchmarks provide limited coverage of such requests. To address this gap, we introduce SkillReason-Bench, a large-scale cross-domain benchmark containing 3,729 queries and a retrieval corpus of 61,228 skills spanning nine domains. We further propose SkillReason, a two-stage framework that uses chain-of-thought reasoning as training-time supervision for skill retrieval. In Stage I, capability reasoning traces generated by a stronger teacher provide explicit supervision through contrastive learning, retrieval distribution alignment, and language modeling, encouraging the retriever to internalize capability reasoning in its query representation. In Stage II, a retrieval-guided GRPO objective encourages the model to explore reasoning trajectories better suited to its own capabilities and more effective for retrieval. At inference, SkillReason directly encodes the original query without autoregressive CoT generation, preserving efficient query-only retrieval. Extensive experiments on SkillReason-Bench, SkillRet, and SRA-Bench show that SkillReason achieves state-of-the-art performance across all three benchmarks, demonstrating that reasoning-enhanced training better bridges the semantic gap between high-level task goals and skill capabilities.
\end{abstract}
\section{Introduction}

Large language models are gradually evolving into agentic systems that interact with external tools and environments to solve complex tasks~\citep{wang2024surveyagents,yao2023react,
schick2023toolformer}. As agents are applied to a wider range of open-ended tasks, relying only on knowledge stored in model parameters is insufficient for stable and scalable task completion. Recent agent systems, such as Claude Code and OpenClaw, therefore introduce reusable external capability units called skills. A skill typically encapsulates task-specific instructions, execution procedures, and auxiliary resources, allowing agents to reuse existing experience and programs rather than solve each similar task from scratch using only the model's internal knowledge~\citep{jiang2026xskill,wang2026webxskill,xia2026metaclaw,li2026agentskillos}. 

\begin{figure}[t]
    \centering
    \includegraphics[width=\columnwidth]{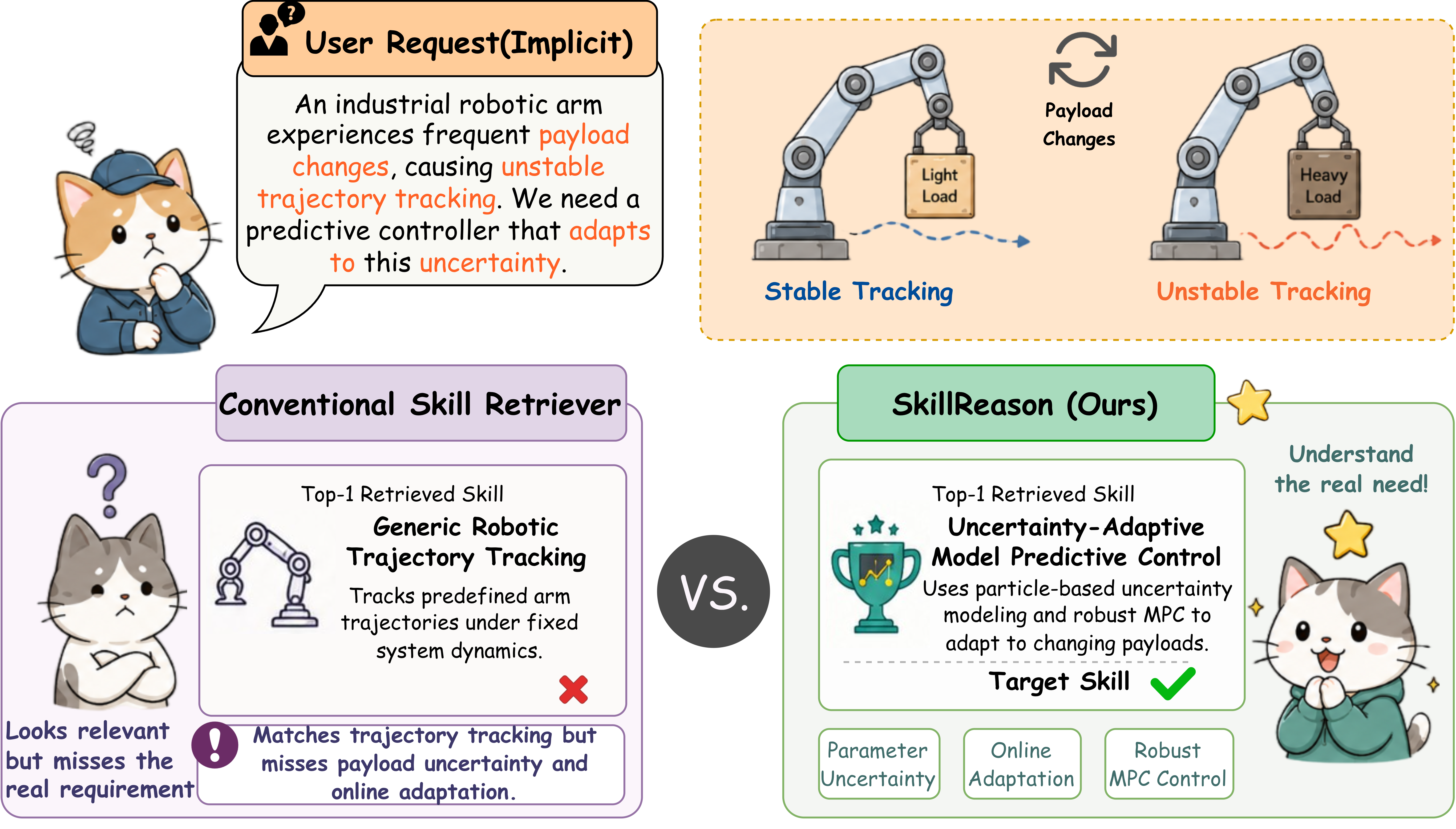}
    \caption{Illustration of skill retrieval from an implicit user request. A conventional retriever may select a superficially related skill, whereas SkillReason infers the underlying capability requirements to retrieve the appropriate skill.}
    \label{fig:introduction}
\end{figure}

However, as reusable skill libraries continue to grow, accurately selecting the appropriate skill from a large candidate pool becomes increasingly challenging~\citep{skillrouter2026}. This challenge stems not only from the growing number of candidates, but more fundamentally from the implicit mapping between user requests and target skills~\citep{chen2024reinvoke,kachuee2025improving,fang2026toolqp}. Realistic requests are often concise and natural, typically stating only the task goal without explaining how it should be accomplished. The retriever must therefore infer the underlying capability requirements from the stated task goal. For example, a user may ask how to maintain stable trajectory tracking for an industrial robotic arm whose payload changes frequently. A surface-level retriever may return a generic trajectory-tracking skill, while failing to infer the need for payload-aware dynamics modeling and uncertainty-adaptive model predictive control.

Existing skill retrieval benchmarks provide useful evaluation settings~\citep{skillret2026}, but they largely focus on queries with explicit procedural descriptions and domain-specific terminology, offering limited coverage of concise and underspecified requests common in realistic agent use. To address this gap, we construct SkillReason-Bench, a large-scale cross-domain skill retrieval benchmark that better reflects natural agent-user interactions. It is designed to evaluate whether retrieval models can infer the underlying capability needs from concise and underspecified task requests, and select the corresponding target skills from a large-scale skill pool.

Conventional contrastive learning for skill retrieval mainly uses relevance signals to directly optimize the matching between requests and skills, providing limited explicit supervision for the semantic reasoning required to infer capability needs from task goals. To address this supervision gap, we propose \textbf{SkillReason}, a two-stage framework that uses chain-of-thought (CoT) reasoning as training-time supervision. In Stage I, capability reasoning traces generated by a stronger teacher model provide explicit supervision that encourages the retriever to internalize capability reasoning in its query representation. However, teacher-generated reasoning traces may not align well with a smaller model's reasoning capacity or the retrieval objective. Stage II therefore introduces a retrieval-guided GRPO objective that encourages the model to explore reasoning trajectories better suited to its own capabilities and more effective for retrieval, thereby further strengthening its semantic reasoning ability. Together, these stages enable the retriever to better infer implicit capability requirements from concise, underspecified requests while preserving efficient query-only inference without autoregressive CoT generation.

Our main contributions can be summarized as follows:

\begin{itemize}
    \item We introduce \textbf{SkillReason-Bench}, a large-scale cross-domain skill retrieval benchmark for concise and underspecified user requests. It contains 3{,}729 high-quality test queries and a retrieval pool of 61{,}228 skills.
    \item We propose \textbf{SkillReason}, a two-stage reasoning-enhanced framework for agent skill retrieval from implicit user requests. It introduces chain-of-thought supervision during training to strengthen dense retrievers' ability to infer implicit capability requirements while preserving efficient query-only inference.

   \item Extensive experiments on SkillReason-Bench, SkillRet, and SRA-Bench show that SkillReason achieves state-of-the-art performance on all three benchmarks, demonstrating that reasoning-enhanced training can better bridge the semantic gap between high-level task goals and skill capabilities.
\end{itemize}
\section{Related Work}
\label{sec:related-work}

\paragraph{Agent Skills.}
LLM agents increasingly extend their parametric capabilities with external
knowledge, tools, and reusable skills~\citep{liu2026skillswild,wang2026skillx,shen2026skillfoundry,zhang2026coevoskills,shi2025autotools,gao2025multimodalagenttuning,lyu2025adapting,liu2026graphofskills}. Unlike retrieved knowledge passages or
individual API calls, a skill packages task-specific instructions, execution
procedures, and auxiliary resources into a reusable capability unit. This
allows agents to reuse structured execution knowledge across related tasks,
but also creates a new selection problem as skill libraries continue to grow.

\paragraph{Skill Retrieval.}
Recent work formulates skill selection as a retrieval problem.
SkillRouter~\citep{skillrouter2026} proposes a retrieve--rerank pipeline for
large and highly overlapping skill libraries, using full skill documents for
both retrieval and reranking. SkillRet~\citep{skillret2026} constructs a
large-scale benchmark with disjoint training and evaluation skill pools, while
SRA-Bench~\citep{srabench2026} evaluates skill retrieval together with downstream skill
incorporation and application. These studies establish
useful retrieval architectures and evaluation settings, but their queries often
contain explicit procedural descriptions, domain terminology, or functional
cues. In contrast, realistic users frequently state only the desired outcome.
SkillReason-Bench therefore focuses on concise and underspecified requests,
requiring retrievers to infer the capabilities needed to identify the target
skill.

\paragraph{Reasoning-Enhanced Retrieval.}
Recent work has explored using model-generated reasoning to improve dense
retrieval~\citep{liu2025rite,yan2026o1embedder}. RQR learns to reformulate reasoning-intensive queries through
reinforcement learning~\citep{qin2025reinforced}, while Think-Then-Embed and
LREM generate reasoning traces before computing query
representations~\citep{cui2025think,tang2025large}. These studies show that
explicit reasoning can help bridge the semantic gap between user queries and
relevant documents. However, they rely on autoregressive reasoning generation
at inference time before retrieval. In contrast, SkillReason explicitly optimizes the raw-query retrieval path and transfers ranking preferences from a privileged reasoning-augmented view, enabling direct query-only retrieval without autoregressive generation at inference.
\section{SkillReason-Bench}
\label{sec:skillreason-bench}

\begin{figure*}[t]
    \centering
    \includegraphics[width=0.95\textwidth]{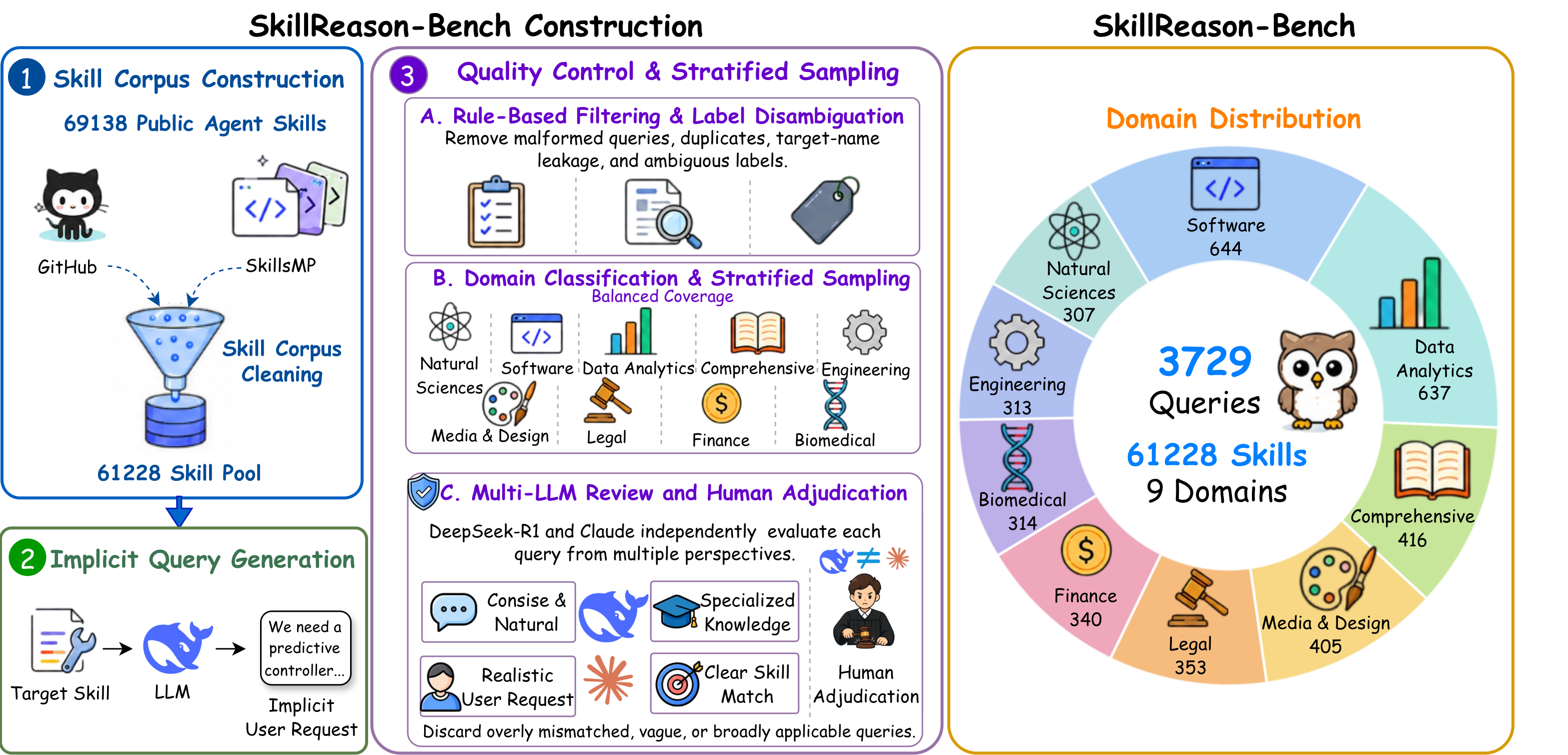}
    \caption{Overview of the SkillReason-Bench construction pipeline and its domain distribution.}
    \label{fig:benchmark-construction}
\end{figure*}

SkillReason-Bench is a large-scale cross-domain benchmark for
skill retrieval from concise and underspecified user requests. It evaluates
retrieval in a setting commonly observed in real-world agent--user
interactions, where users state what they want to accomplish without explicitly
naming the required skill or describing the full execution procedure. The
benchmark contains 3{,}729 high-quality queries, each constructed from
a target skill, and a retrieval corpus of 61{,}228 skills spanning nine domains,
ranging from software and data analytics to finance, media
and the natural sciences. Figure~\ref{fig:benchmark-construction} presents an overview of
the construction pipeline and the domain distribution of the benchmark.

\paragraph{Skill corpus construction.}
We collect 69{,}138 unique public agent skills from GitHub repositories indexed
by SkillsMP, retaining each skill's name, description, full document, and
repository metadata. We remove records with missing or content-deficient
documents, placeholder or template text, malformed content, and duplicate skill
identifiers, resulting in a retrieval corpus of 61{,}228 skills. We do not perform semantic deduplication or merge skills with similar
functionality, so the resulting corpus retains the functional overlap naturally
present in open skill ecosystems.
\paragraph{Implicit query generation.}
We first use DeepSeek-R1 to select anchor skills with concrete, specialized, and executable capabilities. For each anchor, the LLM generates a concise and
natural request that states what the user wants to accomplish. Natural domain terminology and task-specific details are preserved, while
overly explicit references to the target skill, its core capabilities, or the
full execution procedure are avoided. This design makes the evaluation queries
more closely resemble requests commonly seen in real-world agent--user
interactions, where users state the desired outcome without fully specifying
the execution process. Such queries require the retriever to infer the
capabilities needed to identify the target skill.

\noindent\textbf{Quality control and finalization.}
To ensure the reliability and coverage of SkillReason-Bench, we apply
a multi-stage filtering pipeline. We first use rule-based checks to remove
malformed outputs, invalid targets, queries with abnormal lengths, exact
duplicates, and cases that directly reveal the target skill. We then remove near-duplicate queries using BM25 retrieval,
dense embedding similarity, and lexical overlap. To ensure label stability, we retrieve potentially confusable skills from the full corpus using document-overlap heuristics and dense retrieval, and use DeepSeek-R1 to retain only queries whose annotated targets remain clearly distinguishable from these candidates. Next, we perform domain-stratified sampling to reduce
the dominance of frequent categories and preserve diverse coverage across all
nine domains. DeepSeek-R1 and Claude Haiku 4.5 then independently review each
sampled query for naturalness, informativeness, specialized capability
requirements, and consistency with the target skill. Their decisions are
compared, and disagreements are resolved through human adjudication. We retain
queries that are concise and natural, resemble realistic user requests, require
specialized knowledge or capabilities, and can be clearly matched to the target
skill, while discarding queries that are overly simple, vague, broadly
applicable, or mismatched. The resulting benchmark contains 3{,}729
high-quality queries, each paired with a target skill. We further assess query quality using multiple independent LLM judges,
with detailed results reported in the Appendix.

\begin{figure*}[t]
    \centering
    \includegraphics[width=0.95\textwidth]{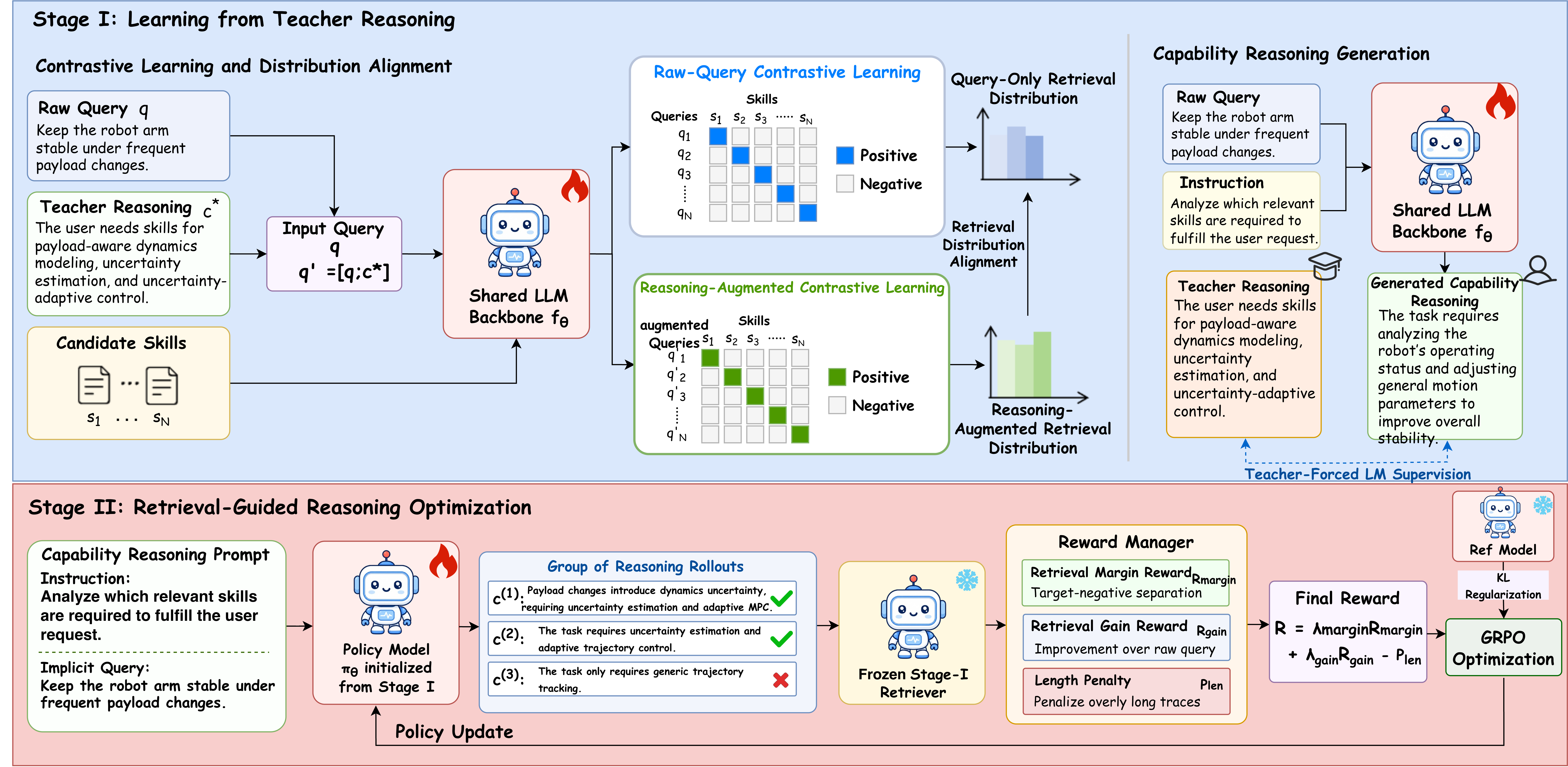}
    \caption{Overview of the SkillReason framework. SkillReason uses capability reasoning
    as privileged supervision in Stage I and retrieval-guided reinforcement
    learning in Stage II to improve the matching between implicit
user requests and target skills. At inference, it requires only the original
query.}
    \label{fig:method}
\end{figure*}

\section{Method}
\label{sec:method}
\subsection{Problem Formulation}
\label{sec:problem-formulation}

Given a user request $q$ and a skill corpus
$\mathcal{S}=\{s_1,\ldots,s_M\}$, the goal is to retrieve the skills most
relevant to completing the requested task. SkillReason uses a shared encoder
$f_\theta$ to map the query and each skill document into normalized
representations. Their relevance score is computed by cosine similarity:
\begin{equation}
h_\theta(q,s)=e_\theta(q)^\top e_\theta(s),
\end{equation}
where $e_\theta(\cdot)$ is obtained by applying $\ell_2$ normalization to
the final-token hidden state of $f_\theta$. At inference, skill
representations are precomputed offline, and the model encodes the original
query once to retrieve the top-ranked skills without generating CoT.
\subsection{Stage I: Learning from Teacher Reasoning}
\label{sec:stage1}

Stage I introduces teacher-generated capability reasoning traces as privileged
supervision. Given a query $q$ and its reasoning trace $c^*$, we construct an
augmented query $\tilde q=[q;c^*]$. We optimize retrieval from both the
original and augmented queries, and align their retrieval distributions to
transfer the capability information in $c^*$ to the query-only representation.
\paragraph{Contrastive learning.}
For each training query $q$, let $s_q^+$ denote its target skill and
$\mathcal{C}_q$ the corresponding candidate set. We apply contrastive learning
to the original query:
\begin{equation}
\mathcal{L}_{\mathrm{raw}}^{\mathrm{CL}}
=
-\log
\frac{\exp(h_\theta(q,s_q^+)/\tau)}
{\sum_{s\in\mathcal{C}_q}\exp(h_\theta(q,s)/\tau)}.
\label{eq:raw-retrieval-loss}
\end{equation}
This objective directly optimizes the query-only retrieval path used at
inference. We then apply the same objective to the augmented query:
\begin{equation}
\mathcal{L}_{\mathrm{cot}}^{\mathrm{CL}}
=
-\log
\frac{\exp(h_\theta(\tilde q,s_q^+)/\tau)}
{\sum_{s\in\mathcal{C}_q}\exp(h_\theta(\tilde q,s)/\tau)}.
\label{eq:cot-retrieval-loss}
\end{equation}
This objective encourages the capability information in $c^*$ to improve the
matching between the query and its target skill.

\paragraph{Retrieval distribution alignment.}
The two contrastive objectives optimize the original and augmented queries
independently, but do not ensure that the original query benefits from the
capability reasoning in $c^*$. We therefore transfer the ranking distribution
of the augmented query to the original query. For an input view $x$, we define
a softened distribution over the candidate skills as
\begin{equation}
p_\theta^T(s\mid x)
=
\frac{\exp\!\left(h_\theta(x,s)/(\tau T)\right)}
{\sum_{s'\in\mathcal{C}_q}
\exp\!\left(h_\theta(x,s')/(\tau T)\right)},
\label{eq:soft-retrieval-distribution}
\end{equation}
where $T$ is the distillation temperature. We use the reasoning-augmented query $\tilde q$ as the privileged view and
the original query $q$ as the query-only view, and minimize
\begin{equation}
\mathcal{L}_{\mathrm{KL}}
=
T^2D_{\mathrm{KL}}\!\left(
\operatorname{sg}\!\left[p_\theta^T(\cdot\mid\tilde q)\right]
\,\middle\|\,
p_\theta^T(\cdot\mid q)
\right),
\label{eq:distribution-transfer}
\end{equation}
where $\operatorname{sg}$ denotes stop-gradient. This objective transfers the relative ranking preferences of the privileged
view to the query-only view, allowing the original query to make similar
retrieval decisions without access to $c^*$ at inference.
\paragraph{Capability reasoning generation.}
We further train the shared backbone to generate the teacher reasoning trace
$c^*$ through teacher forcing, encouraging it to internalize capability
reasoning and initializing the policy for Stage II. Let $\pi_\theta$ denote
the autoregressive generation policy built on $f_\theta$. The objective is
\begin{equation}
\mathcal{L}_{\mathrm{LM}}
=
-\sum_{t=1}^{|c^*|}
\log \pi_\theta(c_t^*\mid q,c_{<t}^*).
\label{eq:cot-lm-loss}
\end{equation}
The query serves as the conditioning context, and the loss is computed only
over the tokens in $c^*$.
The complete Stage-I objective is
\begin{equation}
\begin{aligned}
\mathcal{L}_{\mathrm{SFT}}
={}&\mathcal{L}_{\mathrm{raw}}^{\mathrm{CL}}
+\lambda_{\mathrm{cot}}\mathcal{L}_{\mathrm{cot}}^{\mathrm{CL}}
+\lambda_{\mathrm{KL}}\mathcal{L}_{\mathrm{KL}}\\
&+\lambda_{\mathrm{LM}}\mathcal{L}_{\mathrm{LM}}.
\end{aligned}
\label{eq:stage1-objective}
\end{equation}
Overall, Stage I enables the retriever to internalize capability reasoning in
its query representation and initializes a stable reasoning policy for Stage II.

\subsection{Stage II: Retrieval-Guided Reasoning Optimization}
\label{sec:stage2}
Stage I learns from a single fixed teacher-generated reasoning trace for each
query. However, such a fixed trace may be misaligned with the model's own
reasoning capacity and suboptimal for the retrieval objective. Stage II therefore introduces
retrieval-guided GRPO, which encourages the model to explore reasoning
trajectories better suited to its own capabilities and more effective for
retrieval.

Given an implicit query $q$, the trainable Stage-II policy $\pi_\theta$,
initialized from Stage I, samples $G$ capability reasoning traces
$\{c^{(g)}\}_{g=1}^{G}$. A frozen Stage-I retriever $h_\phi$ encodes the
query and candidate skills to compute the retrieval rewards.

\paragraph{Retrieval reward.}
For an input $x$, we define its retrieval margin as the similarity to the
target skill minus that to the strongest negative:
\begin{equation}
m(x)
=
h_\phi(x,s_q^+)
-
\max_{s^-\in\mathcal{N}_q} h_\phi(x,s^-),
\label{eq:retrieval-margin}
\end{equation}
where $\mathcal{N}_q$ denotes the negative skill set. A larger margin
indicates better separation between the target skill and the negatives.
For each sampled trace $c^{(g)}$, we construct the reasoning-augmented query
$\tilde q^{(g)}=[q;c^{(g)}]$. We first measure its retrieval quality using
the margin reward:
\begin{equation}
R_{\mathrm{margin}}^{(g)}
=
\tanh\left(
\frac{m(\tilde q^{(g)})}{\tau_{\mathrm{margin}}}
\right).
\label{eq:margin-reward}
\end{equation}
We further measure the improvement over the original query:
\begin{equation}
R_{\mathrm{gain}}^{(g)}
=
\operatorname{clip}\left(
\frac{m(\tilde q^{(g)})-m(q)}
{\tau_{\mathrm{gain}}},
-1,1
\right).
\label{eq:gain-reward}
\end{equation}
To discourage unnecessarily long reasoning traces, we introduce a capped
length penalty:
\begin{equation}
P_{\mathrm{len}}(c)
=
\min\left(
P_{\max},
\lambda_{\mathrm{len}}
\left[|c|-L_{\mathrm{budget}}\right]_{+}
\right),
\label{eq:length-penalty}
\end{equation}
where $[x]_{+}=\max(x,0)$, $L_{\mathrm{budget}}$ is the reasoning-length
budget, and $P_{\max}$ caps the total penalty. The final reward is then
defined as
\begin{equation}
R^{(g)}
=
\lambda_{\mathrm{margin}}R_{\mathrm{margin}}^{(g)}
+
\lambda_{\mathrm{gain}}R_{\mathrm{gain}}^{(g)}
-
P_{\mathrm{len}}(c^{(g)}).
\label{eq:retrieval-reward}
\end{equation}
We optimize the policy using the standard GRPO objective with group-relative
reward normalization and KL regularization toward the frozen Stage-I reference
policy. The generation policy and query encoder share the backbone
$f_\theta$. GRPO favors reasoning traces whose capability decompositions
improve retrieval under a frozen reward retriever, while its policy gradients
update the shared backbone parameters that also produce query embeddings.
This parameter sharing allows retrieval-relevant capability reasoning learned
through generation to be internalized in query-only representations.
\section{Experiments}
\label{sec:experiments}

\subsection{Experimental Setup}
\label{sec:exp-setup}
\paragraph{Benchmarks.}

We evaluate SkillReason on three benchmarks that cover different skill
retrieval settings. SkillReason-Bench focuses on concise and underspecified requests whose required capabilities are largely implicit, emphasizing capability inference beyond lexical matching. SkillRet~\cite{skillret2026} features substantially longer and often multi-skill requests with detailed task descriptions and workflow information, primarily testing generalization to unseen skills.
SRA-Bench~\cite{srabench2026} contains 5{,}400 capability-intensive task
instances drawn from six established reasoning, tool-use, medical, mathematical,
and coding benchmarks. It supports decomposed evaluation of the full skill retrieval augmentation pipeline, covering skill retrieval and end-task execution.
\begin{table}[t]
\centering
\footnotesize
\setlength{\tabcolsep}{3.5pt}
\begin{tabular}{@{}lrrr@{}}
\toprule
\textbf{Benchmark} & \textbf{Queries} & \textbf{Skills} & \textbf{Avg. Len.} \\
\midrule
SkillReason-Bench & 3{,}729 & 61{,}228 & 33.2 \\
SkillRet          & 4{,}997 &  6{,}660 & 172.9 \\
SRA-Bench         & 5{,}400 & 26{,}262 & 52.3$^\dagger$ \\
\bottomrule
\end{tabular}
\caption{Benchmark statistics. Skills is the retrieval corpus size and Avg.
Len. is the mean query length in words. $^\dagger$ The average query
length for SRA-Bench is computed excluding MedCalc's long clinical records.}
\label{tab:benchmark-overview}
\end{table}

\setcounter{dbltopnumber}{2}
\renewcommand{\dbltopfraction}{0.95}
\begin{table*}[t]
\centering
\scriptsize
\setlength{\tabcolsep}{2.6pt}
\resizebox{\textwidth}{!}{%
\begin{tabular}{@{}lc cccc cccc cccc@{}}
\toprule
& & \multicolumn{4}{c}{\textbf{SkillReason-Bench}}
& \multicolumn{4}{c}{\textbf{SRA-Bench}}
& \multicolumn{4}{c}{\textbf{SkillRet}} \\
\cmidrule(lr){3-6}\cmidrule(lr){7-10}\cmidrule(lr){11-14}
\textbf{Method} & \textbf{Params.}
& \textbf{R@1} & \textbf{R@5} & \textbf{R@10} & \textbf{N@10}
& \textbf{R@1} & \textbf{R@5} & \textbf{R@10} & \textbf{N@10}
& \textbf{R@1} & \textbf{R@5} & \textbf{R@10} & \textbf{N@10} \\
\midrule
BM25 & --
& 21.29 & 32.66 & 37.38 & 28.86
& 36.69 & 58.47 & 67.35 & 53.92
& 32.83 & 51.09 & 57.68 & 49.78 \\
Qwen3-Embedding-0.6B~\cite{qwen3_embedding} & 0.6B
& 29.90 & 46.53 & 53.69 & 41.00
& 41.34 & 57.08 & 63.76 & 54.46
& 40.95 & 58.27 & 63.74 & 57.18 \\
Qwen3-Embedding-4B~\cite{qwen3_embedding} & 4B
& 34.67 & 54.71 & 62.59 & 47.77
& 46.52 & 66.79 & 73.25 & 62.59
& 36.33 & 55.05 & 61.43 & 53.29 \\
harrier-oss-v1-0.6B~\cite{huang2026harrier} & 0.6B
& 36.55 & 52.91 & 59.43 & 47.28
& 48.48 & 68.52 & 76.22 & 64.88
& 47.22 & 67.09 & 73.66 & 66.82 \\
jina-embeddings-v5-text-small~\cite{akram2026jina} & 0.6B
& 36.02 & 52.70 & 59.56 & 47.18
& 43.02 & 64.47 & 72.22 & 60.08
& 42.77 & 59.76 & 65.74 & 59.74 \\
Qwen3-Embedding-8B~\cite{qwen3_embedding} & 8B
& 37.76 & 58.89 & 65.97 & 51.34
& 50.18 & 74.54 & 82.11 & 69.72
& 42.35 & 60.78 & 66.47 & 59.66 \\
Octen-Embedding-8B~\cite{octen2025rteb} & 8B
& 50.47 & 70.18 & 76.32 & 63.05
& 54.09 & 75.72 & 83.39 & 71.27
& 48.40 & 65.67 & 71.42 & 65.91 \\
KaLM-Gemma3-12B~\cite{zhao2025kalm} & 12B
& 37.22 & 54.55 & 60.96 & 48.59
& 50.00 & 73.41 & 81.63 & 68.95
& 35.32 & 55.34 & 62.83 & 53.39 \\
SkillRouter-Embedding-0.6B~\cite{skillrouter2026} & 0.6B
& 28.00 & 43.12 & 49.85 & 38.18
& 32.08 & 52.95 & 61.24 & 47.84
& 51.74 & 70.88 & 75.90 & 70.69 \\
SkillRet-Embedding-0.6B~\cite{skillret2026} & 0.6B
& 26.95 & 40.79 & 47.12 & 36.30
& 41.81 & 62.20 & 68.82 & 57.77
& 44.64 & 68.83 & 75.75 & 66.81 \\
\midrule
\textbf{SkillReason-0.6B (ours)} & 0.6B
& 45.16 & 62.43 & 69.00 & 56.49
& 52.81 & 79.26 & 88.18 & 73.58
& \textbf{52.45} & 70.87 & 76.70 & 71.03 \\
\textbf{SkillReason-4B (ours)} & 4B
& \textbf{53.58} & \textbf{72.70} & \textbf{78.84} & \textbf{65.93}
& \textbf{60.90} & \textbf{85.07} & \textbf{91.56} & \textbf{80.28}
& 52.03 & \textbf{71.94} & \textbf{77.93} & \textbf{71.59} \\
\bottomrule
\end{tabular}%
}
\caption{Retriever performance (\%) across three skill retrieval
benchmarks. R and N denote Recall and NDCG. All results are reproduced under the same experimental setup with a maximum context length of 4,096 tokens. Best result per metric is bolded.}
\label{tab:main-results}
\end{table*}

\begin{table*}[t]
\centering
\scriptsize
\setlength{\tabcolsep}{2.8pt}
\resizebox{\textwidth}{!}{%
\begin{tabular}{@{}l cccc cccc cccc@{}}
\toprule
& \multicolumn{4}{c}{\textbf{SkillReason-Bench}}
& \multicolumn{4}{c}{\textbf{SRA-Bench}}
& \multicolumn{4}{c}{\textbf{SkillRet}} \\
\cmidrule(lr){2-5}\cmidrule(lr){6-9}\cmidrule(lr){10-13}
\textbf{Configuration}
& \textbf{R@1} & \textbf{R@5} & \textbf{R@10} & \textbf{N@10}
& \textbf{R@1} & \textbf{R@5} & \textbf{R@10} & \textbf{N@10}
& \textbf{R@1} & \textbf{R@5} & \textbf{R@10} & \textbf{N@10} \\
\midrule
Base model
& 29.90 & 46.53 & 53.69 & 41.00
& 41.34 & 57.08 & 63.76 & 54.46
& 40.95 & 58.27 & 63.74 & 57.18 \\
\midrule
\multicolumn{13}{@{}l}{\textit{Stage I: Supervised fine-tuning}} \\
$+$ Raw CL
& 35.18 & 50.36 & 55.94 & 45.07
& 50.04 & 74.20 & 83.17 & 69.29
& 43.63 & 63.45 & 69.97 & 62.29 \\
$+$ CoT CL
& 39.88 & 55.78 & 62.48 & 50.48
& 51.79 & 76.14 & 85.56 & 71.22
& \textbf{53.81} & \textbf{73.55} & 79.15 & 73.30 \\
$+$ Distribution Transfer
& 41.81 & 57.42 & 64.20 & 52.33
& 52.62 & 77.39 & 86.80 & 72.36
& 53.64 & 73.47 & 79.28 & 73.21 \\
$+$ CoT LM
& 42.29 & 58.19 & 65.08 & 53.02
& 52.82 & 77.72 & 86.95 & 72.68
& 53.56 & 73.49 & \textbf{79.32} & \textbf{73.35} \\
\midrule
\multicolumn{13}{@{}l}{\textit{Stage II: Reinforcement learning}} \\
$+$ Retrieval-Guided GRPO 
& \textbf{45.16} & \textbf{62.43} & 69.00 & \textbf{56.49}
& 52.81 & 79.26 & \textbf{88.18} & 73.58
& 52.45 & 70.87 & 76.70 & 71.03 \\
\midrule
\multicolumn{13}{@{}l}{\textit{Inference-Time CoT Augmentation}} \\
$+$ Self-generated CoT augmentation
& 44.73 & 62.24 & \textbf{69.08} & 56.08
& \textbf{52.90} & \textbf{80.18} & 87.78 & \textbf{73.92}
& 51.35 & 70.16 & 76.24 & 70.14 \\
\bottomrule
\end{tabular}
}
\caption{Detailed ablation results (\%) using Qwen3-Embedding-0.6B as the
base model. Stage-I and Stage-II components are added cumulatively, while the
final row evaluates self-generated CoT augmentation using the same Stage-II
checkpoint. R and N denote Recall and NDCG.}
\label{tab:two-stage-ablation}
\end{table*}

\begin{table*}[t]
\centering
\scriptsize
\setlength{\tabcolsep}{2.2pt}
\resizebox{\textwidth}{!}{%
\begin{tabular}{@{}l cccc cccc cccc@{}}
\toprule
& \multicolumn{4}{c}{\textbf{SkillReason-Bench}}
& \multicolumn{4}{c}{\textbf{SRA-Bench}}
& \multicolumn{4}{c}{\textbf{SkillRet}} \\
\cmidrule(lr){2-5}\cmidrule(lr){6-9}\cmidrule(lr){10-13}
\textbf{Pipeline}
& \textbf{R@1} & \textbf{R@5} & \textbf{R@10} & \textbf{N@10}
& \textbf{R@1} & \textbf{R@5} & \textbf{R@10} & \textbf{N@10}
& \textbf{R@1} & \textbf{R@5} & \textbf{R@10} & \textbf{N@10}
\\
\midrule
Qwen3-Emb-0.6B $\times$ Qwen3-Rank-0.6B
& 39.98 & 54.63 & 58.03 & 48.94
& 50.95 & 67.06 & 69.75 & 64.33
& 46.56 & 64.41 & 67.48 & 63.50 \\
Qwen3-Emb-4B $\times$ Qwen3-Rank-4B
& 48.89 & 65.14 & 68.06 & 58.72
& 55.37 & 73.90 & 78.40 & 70.53
& 46.83 & 63.86 & 66.89 & 63.35 \\
SkillRouter-Emb-0.6B $\times$ SkillRouter-Rank-0.6B
& 43.47 & 53.31 & 55.40 & 49.53
& 49.27 & 62.33 & 65.90 & 59.94
& 56.60 & 77.01 & 79.60 & 76.86 \\
\midrule
SkillReason-Emb-0.6B $\times$ Qwen3-Rank-0.6B
& 46.26 & 65.06 & 71.25 & 58.42
& 57.11 & 83.41 & 90.01 & 77.82
& 49.36 & 73.91 & 79.13 & 71.92 \\
SkillReason-Emb-0.6B $\times$ Qwen3-Rank-4B
& 51.41 & 69.78 & 73.45 & 62.63
& 62.29 & 85.78 & 91.34 & 80.91
& 50.72 & 74.67 & 79.44 & 72.83 \\
SkillReason-Emb-0.6B $\times$ SkillReason-Rank-0.6B
& 51.38 & 68.06 & 72.78 & 61.86
& 61.84 & 85.74 & 90.90 & 80.66
& 56.72 & 78.00 & 80.52 & 77.52 \\
SkillReason-Emb-0.6B $\times$ SkillReason-Rank-4B
& 57.90 & 71.12 & 74.23 & 66.19
& 68.59 & 88.52 & 92.39 & 84.90
& \textbf{59.43} & 79.56 & 80.95 & 79.77 \\
SkillReason-Emb-4B $\times$ SkillReason-Rank-4B
& \textbf{60.93} & \textbf{77.50} & \textbf{81.66} & \textbf{71.34}
& \textbf{69.27} & \textbf{90.30} & \textbf{94.32} & \textbf{86.47}
& 59.39 & \textbf{81.28} & \textbf{82.98} & \textbf{81.02} \\
\bottomrule
\end{tabular}%
}
\caption{Retrieve--rerank pipeline results (\%). Each reranker reorders the
top-20 candidates returned by its paired retriever. R and N denote Recall and NDCG.}
\label{tab:rerank-results}
\end{table*}

\paragraph{Training data.}
We construct the training data from the large-scale skill pool released by
SkillRouter, which is sourced from the Claude Skill Registry. The training queries are disjoint from all evaluation queries, which are
collected from different sources and exhibit distinct distributions. This setting evaluates cross-corpus generalization to independently constructed
queries drawn from distinct distributions. For each skill, we use Claude Sonnet 4.6 to generate a user query and a corresponding
capability reasoning trace. We then apply rule-based checks, embedding-based
deduplication, and DeepSeek-R1 quality filtering to remove invalid, redundant,
or mismatched samples. The retained data are rebalanced across skill
categories, yielding 30K training instances for Stage I. For Stage II, to better match the target setting of implicit user requests, we use Claude Haiku 4.5 to rewrite the Stage-I queries into more implicit formulations while preserving their original task intent and target skills. 
\paragraph{Implementation details.}
We train 0.6B and 4B SkillReason models initialized from Qwen3-Embedding,
using supervised fine-tuning in Stage I and retrieval-guided GRPO with eight
rollouts per query in Stage II. For retrieve--rerank experiments, rerankers
initialized from Qwen3-Reranker reorder the top-20 candidates returned by the
corresponding SkillReason retrievers. Detailed training-data construction procedures, representative examples, and training configurations are provided in the Appendix.

\begin{table*}[t]
\centering
\scriptsize
\setlength{\tabcolsep}{5.0pt}
\resizebox{\textwidth}{!}{%
\begin{tabular}{@{}l ccccc@{}}
\toprule
\textbf{Skill Condition}
& \textbf{Qwen3-4B} & \textbf{Qwen3-32B} & \textbf{Qwen3-235B}
& \textbf{Mistral3.1-24B} & \textbf{Llama-3-70B} \\
\midrule
No skills & 38.81 & 50.75 & 54.30 & 44.17 & 42.02 \\
Gold skills & 64.23 & 68.84 & 70.28 & 65.21 & 61.42 \\
\midrule
Qwen3-Emb-0.6B $\times$ Qwen3-Rank-0.6B
& 58.00 & 63.71 & 64.64 & 58.23 & 55.37 \\
Qwen3-Emb-4B $\times$ Qwen3-Rank-4B
& 59.77 & 63.64 & 66.32 & 60.27 & 55.63 \\
SkillRouter-Emb-0.6B $\times$ SkillRouter-Rank-0.6B
& 58.56 & 63.90 & 64.73 & 59.77 & 53.77 \\
SkillReason-Emb-0.6B $\times$ SkillReason-Rank-0.6B
& 62.28 & 67.87 & 67.54 & 63.03 & 58.79 \\
SkillReason-Emb-4B $\times$ SkillReason-Rank-4B
& \textbf{63.84} & \textbf{69.09} & \textbf{68.96} & \textbf{64.47} & \textbf{59.83} \\
\bottomrule
\end{tabular}%
}
\caption{End-to-end task accuracy (\%) on SRA-Bench. For each
retrieve--rerank pipeline, the top-1 retrieved skill is provided to the
downstream LLM under the same direct-inference setting. Results are
macro-averaged across six evaluation suites.}
\label{tab:end-to-end-results}
\end{table*}
\subsection{Main Results}
\label{sec:main-results}

We compare SkillReason with BM25, general-purpose embedding models, and
existing skill-specific retrievers. We report Recall@1, Recall@5, Recall@10,
and NDCG@10 on all three benchmarks. As shown in Table~\ref{tab:main-results}, SkillReason-4B achieves the best
reported performance on most metrics. SkillReason-0.6B achieves an average absolute
gain of 17.56 points in Recall@10 over its Qwen3-Embedding-0.6B backbone. Despite using only 0.6B parameters, it consistently
outperforms Qwen3-Embedding-8B across all metrics.

The cross-benchmark comparison highlights a key limitation of existing
skill-specific retrievers. As shown in
Table~\ref{tab:benchmark-overview}, the three benchmarks differ
substantially in query length. SkillRouter and SkillRet perform strongly on
SkillRet, whose queries are relatively long and often contain explicit
procedural cues, but their gains narrow on SRA-Bench, and both fall below Qwen3-Embedding-0.6B on
SkillReason-Bench. In contrast, SkillReason remains strong across all three benchmarks. This
suggests that reasoning-enhanced training helps the retriever infer implicit
capability requirements from concise and underspecified user requests, rather
than relying primarily on explicit lexical and procedural cues, thereby improving robustness across queries with varying levels of
detail.
\subsection{Ablation Study}
\label{sec:ablation}

As shown in Table~\ref{tab:two-stage-ablation}, incorporating CoT CL
consistently improves performance across all three benchmarks. The
improvements on SkillReason-Bench and SRA-Bench suggest that CoT supervision
helps the retriever go beyond surface-level query cues and infer the
capabilities required to complete the task, while the improvement on SkillRet
shows that it also benefits longer and more procedurally detailed requests. Adding retrieval distribution alignment further improves
SkillReason-Bench and SRA-Bench, suggesting that transferring ranking
preferences from the reasoning-augmented view helps the query-only
representation make better retrieval decisions when explicit cues are
limited. The CoT language modeling objective  further improves performance by explicitly training the shared
backbone to generate capability reasoning traces, encouraging the reasoning
process to be internalized in the query representation and providing a better
initialization for Stage II.

Stage II with retrieval-guided GRPO further improves performance on SkillReason-Bench and
most metrics on SRA-Bench, suggesting that retrieval feedback enables the
model to move beyond imitating fixed teacher-generated CoTs and explore
reasoning trajectories that better support skill retrieval. It does not further improve SkillRet, whose queries are substantially longer and more procedurally explicit, indicating that the gains from Stage II are concentrated in more implicit-query settings. More importantly,
augmenting the query with self-generated CoT at inference does not yield
consistent gains over query-only retrieval. This suggests that the
retrieval-relevant reasoning acquired during training has been effectively
internalized into the query representation, enabling the model to retain the
benefits of capability reasoning without additional generation at inference
time. 
\subsection{Retrieve--Rerank Pipelines}
\label{sec:rerank-results}
Table~\ref{tab:rerank-results} evaluates complete retrieve--rerank pipelines,
where each reranker reorders the top-20 candidates returned by its paired
retriever. SkillReason-based pipelines achieve strong performance across all
three benchmarks. In particular, the fully 0.6B SkillReason pipeline
outperforms the 4B Qwen3 pipeline on every evaluation metric, showing that
the advantages of SkillReason extend beyond retriever-only evaluation to the
complete retrieval pipeline.
\subsection{End-to-end task evaluation}
\label{sec:end-to-end-results}

To evaluate whether retrieval improvements translate into better downstream
performance, we provide the top-1 skill retrieved by each pipeline to the
downstream LLM and measure task accuracy on SRA-Bench. As shown in
Table~\ref{tab:end-to-end-results}, SkillReason consistently improves
end-to-end accuracy across all five downstream models. Notably, the 0.6B
SkillReason pipeline outperforms the 4B Qwen3 pipeline for every downstream
model. The 4B SkillReason pipeline raises task success rates by between 14.66 and 25.03
percentage points across all five downstream LLMs, closely matching the performance obtained with gold skills. These results
demonstrate that SkillReason’s retrieval gains translate into higher downstream task success rates.
\section{Conclusion}

We introduced SkillReason-Bench, a large-scale benchmark for skill retrieval
from concise and underspecified user requests, and proposed SkillReason, a
two-stage framework that combines teacher-supervised capability reasoning
with retrieval-guided reinforcement learning. Experiments on three retrieval benchmarks and an end-to-end task evaluation
demonstrate substantial improvements in both retrieval quality and downstream
task performance. Our work presents a new approach that distills CoT reasoning into retrieval
representations, retaining the benefits of inference-time CoT augmentation
without introducing additional generation latency.

\bibliography{aaai2027}

\def\SkillReasonMainDocument{1}
\ifdefined\SkillReasonMainDocument
  \def\SkillReasonEndSupplement{}
\else
  \def\SkillReasonEndSupplement{\end{document}}
  \documentclass[letterpaper]{article}
  \usepackage[submission]{aaai2027}
  \nocopyright
  \usepackage[hyphens]{url}
  \usepackage{graphicx}
  \urlstyle{rm}
  \def\UrlFont{\rm}
  \usepackage{natbib}
  \usepackage{caption}
  \frenchspacing

  \usepackage{algorithm}
  \usepackage{algorithmic}
  \usepackage{newfloat}
  \usepackage{listings}
  \usepackage{booktabs}
  \usepackage{amsmath}
  \usepackage{amssymb}
  \usepackage{multirow}
  \usepackage[most]{tcolorbox}
  \usepackage{dblfloatfix}
\fi
\ifdefined\SkillReasonMainDocument
\else
\DeclareCaptionStyle{ruled}{labelfont=normalfont,labelsep=colon,strut=off}
\lstset{
  basicstyle={\footnotesize\ttfamily},
  numbers=left,
  numberstyle=\footnotesize,
  xleftmargin=2em,
  aboveskip=0pt,
  belowskip=0pt,
  showstringspaces=false,
  tabsize=2,
  breaklines=true
}
\floatstyle{ruled}
\newfloat{listing}{tb}{lst}{}
\floatname{listing}{Listing}

\setlength{\dbltextfloatsep}{10pt plus 2pt minus 2pt}
\setlength{\dblfloatsep}{10pt plus 2pt minus 2pt}
\makeatletter
\setlength{\@dblfptop}{0pt}
\setlength{\@dblfpsep}{12pt}
\setlength{\@dblfpbot}{0pt plus 1fil}
\makeatother

\definecolor{promptbg}{HTML}{EEF4FB}
\definecolor{promptframe}{HTML}{5E81AC}
\definecolor{casebg}{HTML}{F5F8FC}
\definecolor{caseframe}{HTML}{7A9BC2}
\newtcblisting{promptbox}[1][]{
  enhanced,
  breakable,
  listing only,
  colback=promptbg,
  colframe=promptframe,
  fonttitle=\bfseries\small,
  title={#1},
  boxrule=0.5pt,
  arc=2pt,
  left=6pt,
  right=6pt,
  top=4pt,
  bottom=4pt,
  listing options={
    basicstyle=\footnotesize\ttfamily,
    numbers=none,
    xleftmargin=0pt,
    aboveskip=0pt,
    belowskip=0pt,
    breaklines=true,
    columns=fullflexible,
    keepspaces=true,
    showstringspaces=false,
    tabsize=2
  }
}
\newtcolorbox{casebox}[1][]{
  enhanced,
  breakable,
  colback=casebg,
  colframe=caseframe,
  fonttitle=\bfseries\small,
  fontupper=\small,
  title={#1},
  boxrule=0.5pt,
  arc=2pt,
  left=6pt,
  right=6pt,
  top=4pt,
  bottom=4pt
}

\newcommand{\todo}[1]{\textbf{[TODO: #1]}}
\fi

\setcounter{secnumdepth}{2}

\ifdefined\SkillReasonMainDocument
  \clearpage
\else
  \pdfinfo{
  /TemplateVersion (2027.1)
  }
  \title{SkillReason: Reasoning-Enhanced Agent Skill Retrieval for Implicit User Requests\\
   Supplementary Material}
  \author{Anonymous Submission}
  \affiliations{}
  \begin{document}
  \maketitle
\fi
\appendix

\section{Benchmark Reliability}
\label{app:benchmark-validation}

\subsection{Independent Multi-LLM Quality Assessment}
\label{app:benchmark-validation-protocol}
\paragraph{Protocol.}
To independently assess the quality of SkillReason-Bench, we draw a
fixed-seed, domain-stratified sample of 300 queries from the finalized
3{,}729-query benchmark. Qwen3-235B-A22B and GPT-5.6-Luna independently
evaluate the sample. Neither model participated in benchmark construction,
which used DeepSeek-R1 and Claude Haiku 4.5. Each judge receives the user
request together with the gold skill's name, description, and document,
truncated to 12{,}000 characters, and performs the evaluation using
deterministic decoding. We assess four criteria on a 1--5 scale:
naturalness, gold-skill alignment, skill necessity, and implicitness.
A sample passes if its naturalness and gold-skill alignment scores are at
least 4, its skill necessity and implicitness scores are at least 3, and no
direct target leakage is detected. We apply a predefined consistency check
to initially disputed cases and manually adjudicate the remaining
cross-model disagreements. The core evaluation prompt and output schema are
provided in Section~\ref{app:benchmark-validation-prompt}.

\paragraph{Results and human adjudication.}
Table~\ref{tab:benchmark-quality-audit} summarizes the independent
evaluations and final adjudicated outcomes. Qwen3-235B-A22B marks 298 of
300 samples as passing (99.3\%), while GPT-5.6-Luna marks 293 as passing
(97.7\%). The two judges jointly accept 293 samples (97.7\%), jointly reject
two (0.7\%), and disagree on the remaining five (1.7\%), yielding agreement
on 295 of 300 samples (98.3\%).

\begin{center}
  \captionof{table}{Independent quality audit on 300 domain-stratified
  samples. Scores are means on a 1--5 scale; leakage and pass are rates.}
  \label{tab:benchmark-quality-audit}
  \small
  \resizebox{0.98\columnwidth}{!}{%
  \begin{tabular}{@{}lrrrrrr@{}}
    \toprule
    \textbf{Judge} & \textbf{Pass (\%)} & \textbf{Natural} & \textbf{Align.}
      & \textbf{Necess.} & \textbf{Implicit} & \textbf{Leakage (\%)} \\
    \midrule
    Qwen3-235B-A22B & 99.33 & 4.997 & 4.930 & 4.677 & 4.727 & 0.00 \\
    GPT-5.6-Luna & 97.67 & 4.970 & 4.847 & 4.347 & 4.940 & 0.33 \\
    \bottomrule
  \end{tabular}%
  }
\end{center}

Human adjudication finds all five disputed cases to be valid under the
capability-level relevance criterion, while confirming mild gold-scope
mismatches in the two cases rejected by both judges. The final adjudicated
pass rate is therefore 298/300 (99.3\%). We also manually inspect 50 randomly
sampled judge rationales and verify that their assessments are consistent
with the corresponding requests and skill documents.

\noindent\textbf{Jointly Rejected Cases.}%
\label{app:benchmark-validation-cases}\par
\vspace*{0.15\baselineskip}

The two samples rejected by both independent judges are examined below.
Human review confirms that both contain mild mismatches between the scope of
the user request and that of the annotated gold skill.

\begin{casebox}[Case 1: Vitest Isolation (UID 24495)]
\textbf{User request.} ``When writing unit tests in Vitest, shared global
variables cause random failures between test cases. How can I configure it so
each test runs in complete isolation?''

\medskip
\textbf{Gold skill.} \texttt{vitest-environment-globals-setups}, which covers
Vitest environment and global-variable setup.

\medskip
\textbf{Human assessment.} The request is natural and requires specialized
testing knowledge. However, the documented skill focuses on environment and
global-variable setup rather than the broader per-test isolation objective.
We therefore classify this item as a mild gold-scope mismatch.
\end{casebox}

\begin{casebox}[Case 2: Visual Offset Auditing (UID 40598)]
\textbf{User request.} ``In my 3D modeling software, when users place objects
on the floor or ceiling, the models often float or become embedded. How can we
automatically check this and generate a remediation report?''

\medskip
\textbf{Gold skill.} \texttt{slabbed-visual-offset-audit}, which focuses on
verifying model and outline alignment on slab tops and undersides.

\medskip
\textbf{Human assessment.} The skill provides a closely related auditing
capability, but its documented scope is narrower than the requested general
automatic inspection and remediation workflow. We therefore classify this
item as a mild gold-scope mismatch.
\end{casebox}

\subsection{Sensitivity Analysis with Validated Alternative Skills}
\label{app:alternative-gold}

\begin{table*}[!t]
  \centering
  \caption{Sensitivity analysis with validated alternative skills on
  SkillReason-Bench (\%). Original recall uses the released benchmark
  annotations, while expanded recall additionally credits candidates validated
  by Qwen3-235B-A22B as valid alternatives for the corresponding query.
  R@1 is computed from the top-ranked result in the same top-5 rankings, and
  $\Delta$ denotes the absolute change after expanding the positive set.}
  \label{tab:strict-alternative-gold-results}
  \scriptsize
  \setlength{\tabcolsep}{5pt}
  \begin{tabular}{@{}lrrrrrr@{}}
    \toprule
    \textbf{Retriever}
      & \textbf{Original R@1} & \textbf{Expanded R@1} & \textbf{$\Delta$R@1}
      & \textbf{Original R@5} & \textbf{Expanded R@5} & \textbf{$\Delta$R@5} \\
    \midrule
    BM25                         & 21.29 & 22.90 & $+1.61$ & 32.66 & 36.15 & $+3.49$ \\
    Qwen3-Embedding-0.6B         & 29.90 & 32.45 & $+2.55$ & 46.53 & 50.90 & $+4.37$ \\
    Qwen3-Embedding-4B           & 34.67 & 37.19 & $+2.52$ & 54.71 & 58.81 & $+4.10$ \\
    Harrier-0.6B                 & 36.55 & 39.13 & $+2.57$ & 52.91 & 56.64 & $+3.73$ \\
    Jina-Embedding-v5-Small      & 36.02 & 38.80 & $+2.79$ & 52.70 & 57.01 & $+4.32$ \\
    Qwen3-Embedding-8B           & 37.76 & 40.28 & $+2.52$ & 58.89 & 63.13 & $+4.24$ \\
    Octen-Embedding-8B           & 50.47 & 54.01 & $+3.54$ & 70.18 & 74.09 & $+3.92$ \\
    KaLM-Gemma3-12B              & 37.22 & 40.49 & $+3.27$ & 54.55 & 58.65 & $+4.10$ \\
    SkillRouter-Embedding-0.6B   & 28.00 & 29.36 & $+1.37$ & 43.12 & 46.07 & $+2.95$ \\
    SkillRet-Embedding-0.6B      & 26.95 & 28.61 & $+1.66$ & 40.79 & 44.41 & $+3.62$ \\
    SkillReason-0.6B    & 45.16 & 47.87 & $+2.71$ & 62.43 & 66.80 & $+4.37$ \\
    SkillReason-4B      & \textbf{53.58} & \textbf{56.23}
      & $\mathbf{+2.65}$ & \textbf{72.70} & \textbf{76.54} & $\mathbf{+3.83}$ \\
    \bottomrule
  \end{tabular}
\end{table*}

\begin{table*}[!t]
  \centering
  \caption{Sensitivity analysis with validated alternative skills on
  SkillRet (\%). Original recall uses the same official evaluation protocol and
  values reported in the main table. Expanded recall additionally credits a
  validated alternative as covering one corresponding uncovered gold skill,
  with each annotated
  gold skill credited at most once. R@1 is computed from the top-ranked result
  in the same top-5 rankings, and $\Delta$ denotes the absolute change after
  expanding the positive set.}
  \label{tab:strict-alternative-gold-skillret}
  \scriptsize
  \setlength{\tabcolsep}{5pt}
  \begin{tabular}{@{}lrrrrrr@{}}
    \toprule
    \textbf{Retriever}
      & \textbf{Original R@1} & \textbf{Expanded R@1} & \textbf{$\Delta$R@1}
      & \textbf{Original R@5} & \textbf{Expanded R@5} & \textbf{$\Delta$R@5} \\
    \midrule
    BM25                         & 32.83 & 35.03 & $+2.20$ & 51.09 & 55.29 & $+4.21$ \\
    Qwen3-Embedding-0.6B         & 40.95 & 43.19 & $+2.24$ & 58.27 & 60.17 & $+1.90$ \\
    Qwen3-Embedding-4B           & 36.33 & 39.28 & $+2.95$ & 55.05 & 58.44 & $+3.38$ \\
    Harrier-0.6B                 & 47.22 & 50.13 & $+2.91$ & 67.09 & 70.78 & $+3.69$ \\
    Jina-Embedding-v5-Small      & 42.77 & 45.02 & $+2.24$ & 59.76 & 62.73 & $+2.97$ \\
    Qwen3-Embedding-8B           & 42.35 & 45.40 & $+3.05$ & 60.78 & 63.79 & $+3.01$ \\
    Octen-Embedding-8B           & 48.40 & 50.96 & $+2.56$ & 65.67 & 68.36 & $+2.69$ \\
    KaLM-Gemma3-12B              & 35.32 & 38.05 & $+2.73$ & 55.34 & 59.56 & $+4.23$ \\
    SkillRouter-Embedding-0.6B   & 51.74 & 53.76 & $+2.02$ & 70.88 & 72.92 & $+2.04$ \\
    SkillRet-Embedding-0.6B      & 44.64 & 49.76 & $+5.12$ & 68.83 & 74.85 & $+6.02$ \\
    SkillReason-0.6B    & \textbf{52.45} & \textbf{55.15}
      & $+2.70$ & 70.87 & 73.36 & $+2.49$ \\
    SkillReason-4B      & 52.03 & 54.99
      & $+2.96$ & \textbf{71.94} & \textbf{75.26} & $+3.32$ \\
    \bottomrule
  \end{tabular}
\end{table*}

\newcommand{\additionalablationtable}{%
\begin{table*}[!t]
  \centering
  \caption{Additional ablations and control experiments on SkillReason-0.6B
  (\%). Stage-I controls add each objective individually to Raw CL rather than
  cumulatively. Stage-II data controls start from the complete Stage-I
  checkpoint, while reward ablations remove one component from the complete
  retrieval-guided RL objective. Teacher-CoT augmentation appends a capability
  rationale generated by Qwen3-235B-A22B at inference.}
  \label{tab:additional-ablation-controls}
  \scriptsize
  \setlength{\tabcolsep}{2.8pt}
  \resizebox{\textwidth}{!}{%
  \begin{tabular}{@{}l cccc cccc cccc@{}}
    \toprule
    & \multicolumn{4}{c}{\textbf{SkillReason-Bench}}
    & \multicolumn{4}{c}{\textbf{SRA-Bench}}
    & \multicolumn{4}{c}{\textbf{SkillRet}} \\
    \cmidrule(lr){2-5}
    \cmidrule(lr){6-9}
    \cmidrule(lr){10-13}
    \textbf{Configuration}
    & \textbf{R@1} & \textbf{R@5} & \textbf{R@10} & \textbf{N@10}
    & \textbf{R@1} & \textbf{R@5} & \textbf{R@10} & \textbf{N@10}
    & \textbf{R@1} & \textbf{R@5} & \textbf{R@10} & \textbf{N@10} \\
    \midrule

    \multicolumn{13}{@{}l}{\textit{Stage-I objective controls}} \\

    Raw CL
    & 35.18 & 50.36 & 55.94 & 45.07
    & 50.04 & 74.20 & 83.17 & 69.29
    & 43.63 & 63.45 & 69.97 & 62.29 \\

    Raw CL $+$ CoT CL
    & 39.88 & 55.78 & 62.48 & 50.48
    & 51.79 & 76.14 & 85.56 & 71.22
    & 53.81 & 73.55 & 79.15 & 73.30 \\

    Raw CL $+$ Distribution Transfer
    & 19.76 & 32.64 & 39.02 & 28.65
    & 39.55 & 59.75 & 67.19 & 55.45
    & 41.81 & 60.57 & 67.02 & 59.05 \\

    Raw CL $+$ CoT LM
    & 40.55 & 57.04 & 63.37 & 51.42
    & 52.36 & 76.92 & 86.46 & 72.03
    & 52.63 & 72.67 & 78.18 & 72.04 \\

    \midrule
    \multicolumn{13}{@{}l}{\textit{Stage-II data and reward controls}} \\
    Complete Stage I
    & 42.29 & 58.19 & 65.08 & 53.02
    & 52.82 & 77.72 & 86.95 & 72.68
    & 53.56 & 73.49 & 79.32 & 73.35 \\

    \quad $+$ Implicit CL
    & 43.88 & 60.65 & 66.73 & 54.18
    & 51.97 & 77.61 & 86.63 & 72.38
    & 44.31 & 62.10 & 68.47 & 60.94 \\

    \quad $+$ Implicit Full SFT
    & 45.68 & 63.22 & 69.62 & 56.51
    & 52.16 & 79.56 & 88.42 & 73.32
    & 48.15 & 66.73 & 72.97 & 65.98 \\

    Full Stage II (Retrieval-Guided RL)
    & 45.16 & 62.43 & 69.00 & 56.49
    & 52.81 & 79.26 & 88.18 & 73.58
    & 52.45 & 70.87 & 76.70 & 71.03 \\

    \quad w/o Margin Reward
    & 44.38 & 61.87 & 68.28 & 55.71
    & 52.71 & 78.18 & 87.36 & 73.09
    & 51.93 & 70.25 & 75.76 & 70.25 \\

    \quad w/o Gain Reward
    & 43.55 & 60.10 & 66.88 & 54.54
    & 52.77 & 78.43 & 87.39 & 73.14
    & 52.02 & 70.48 & 76.22 & 70.81 \\

    \quad w/o Length Penalty
    & 45.13 & 62.40 & 69.18 & 56.48
    & 52.74 & 78.25 & 87.64 & 73.45
    & 51.01 & 68.52 & 74.49 & 68.84 \\

    \midrule
    \multicolumn{13}{@{}l}{\textit{Inference-time Teacher-CoT augmentation}} \\
    Full Stage II $+$ Teacher CoT
    & 44.95 & 62.59 & 69.38 & 56.61
    & 56.23 & 81.17 & 89.68 & 76.39
    & 52.04 & 70.91 & 76.80 & 70.88 \\

    \bottomrule
  \end{tabular}%
  }
\end{table*}
}

Skill retrieval benchmarks are typically evaluated against a finite set of
annotated target skills, although large open skill libraries may contain
multiple functionally substitutable skills. SkillReason-Bench assigns one
official target skill to each query, whereas SkillRet provides multiple
official gold skills. In both settings, a retrieved skill that can independently
satisfy the request may still be counted as incorrect if it is absent from the
official annotations. We therefore assess the sensitivity of the reported results to such
unannotated but valid alternatives, with relevance determined through
independent review rather than retrieval scores or embedding similarity.

For SkillReason-Bench, we pool the top-5 predictions from all 12 retrievers,
yielding 104{,}010 non-gold predictions and 69{,}263 unique
query--candidate pairs after deduplication. Qwen3-235B-A22B reviews each
candidate together with the user request and the official target skill. The
review assesses whether the candidate provides the core capability required
by the request, covers its key requirements, and can serve as a valid
alternative to the official target. Validated alternatives are treated as
additional positives for the corresponding query. At each cutoff, retrieval is
counted as successful if the ranked results contain either the official target
or one of these validated alternatives. We then rescore the existing top-5
rankings under the expanded positive set and compute R@1 from the top-ranked
result.

After adding the validated alternatives, R@1 increases by between 1.37 and
3.54 percentage points, while R@5 increases by between 2.95 and 4.37
percentage points across the evaluated retrievers. The overall conclusions
remain unchanged: SkillReason-4B achieves the best performance at both
cutoffs, with 56.23 R@1 and 76.54 R@5, while SkillReason-0.6B remains
substantially ahead of the other 0.6B skill-specific retrievers.

We also apply the same candidate-review procedure to SkillRet, which provides
multiple official gold skills for each query. Following its native
fractional-recall protocol, a validated alternative is counted as covering one
corresponding uncovered gold skill, and each official gold skill can receive
credit at most once. Table~\ref{tab:strict-alternative-gold-skillret} reports
the resulting scores.

The relative ordering of the methods remains stable on SkillRet:
SkillReason-0.6B achieves the best R@1, while SkillReason-4B achieves the
best R@5. The same conclusion holds on SkillReason-Bench, where
SkillReason-4B remains best at both cutoffs. Although adding validated
alternatives changes the absolute recall scores, it does not alter the main
comparisons among retrievers on either benchmark. This stability is
particularly important for SkillReason-Bench, whose retrieval corpus contains
61{,}228 candidate skills, approximately $9.2\times$ more than SkillRet.
These results indicate that the evaluation on SkillReason-Bench reliably
preserves the relative performance of different retrieval methods, even with
a substantially larger retrieval corpus.

\additionalablationtable

\section{Additional Experimental Results}
\label{app:additional-results}

This section provides additional experimental analyses that complement the
main results. We first report isolated Stage-I objective controls, matched
Stage-II adaptation comparisons, reward-component ablations, and
inference-time Teacher-CoT augmentation. We then evaluate retrieval on
SkillBench, with particular attention to the complete coverage of multi-skill
requests, and finally analyze the online latency of retrieval, reranking, and
explicit CoT generation. Unless otherwise noted, all evaluations use the same
retrieval corpora and query instructions as the main experiments. 

\subsection{Stage-I objective controls}
The first block of Table~\ref{tab:additional-ablation-controls} isolates the
individual Stage-I objectives by adding each of them directly to Raw CL.
Both CoT CL and CoT LM consistently outperform the Raw CL baseline across
all three benchmarks, showing that capability reasoning supervision is
effective under both contrastive and generative training objectives. CoT LM
is more effective on SkillReason-Bench and SRA-Bench, whereas CoT CL performs
better on SkillRet. This difference suggests that the two objectives provide
complementary supervision: CoT CL directly improves the alignment between
queries and relevant skills, while CoT LM encourages the encoder to model the
intermediate capabilities implied by the request.

Distribution Transfer, in contrast, performs poorly when applied directly to
Raw CL. This suggests that matching retrieval distributions is not a reliable
standalone learning objective and is better used as an auxiliary alignment
signal after meaningful capability representations have been established.
The complete Stage-I model provides the strongest overall balance across the
three benchmarks, indicating that combining the CoT-based objectives with
distribution transfer retains their complementary benefits while avoiding the
instability of distribution matching in isolation.

\subsection{Stage-II data and reward controls}
The second block compares different ways of adapting the complete Stage-I
model on the same implicit-query training data. Continued contrastive learning
and full SFT both improve retrieval on SkillReason-Bench and provide moderate
gains at deeper cutoffs on SRA-Bench. However, both methods substantially
degrade performance on SkillRet. This result shows that additional training
on implicit queries alone does not guarantee robust improvement and may
over-specialize the retriever to the implicit-query training distribution.

Retrieval-guided RL produces a more balanced outcome. It preserves most of
the gains on SkillReason-Bench while maintaining substantially stronger
performance on SkillRet than either implicit CL or implicit full SFT. Its
performance on SRA-Bench also remains stable. The advantage of Stage II
therefore cannot be attributed solely to exposure to additional implicit
queries. Instead, directly optimizing retrieval quality helps the model adapt
to implicit requests without sacrificing as much cross-benchmark
generalization.

The reward ablations further clarify the contribution of each component.
Removing either the margin reward or the gain reward consistently reduces
retrieval quality, confirming that both positive-negative separation and the
retrieval improvement induced by the generated rationale are important. The
gain reward has the clearest effect on SkillReason-Bench, where reasoning over
implicit requests is most central to the task. Removing the length penalty has
only a minor effect on SkillReason-Bench but produces clearer degradation on
SRA-Bench and SkillRet. This pattern suggests that the length penalty mainly
serves as a regularizer, preventing unnecessarily long or overly specialized
reasoning from weakening transfer to other query distributions.

\subsection{Inference-time Teacher-CoT augmentation}
The final block examines whether the query-only SkillReason-0.6B model has
effectively internalized the capability reasoning learned during training.
We use the substantially larger Qwen3-235B-A22B model to generate an explicit
capability rationale for each test query and append it to the original query
at inference. Despite access to these high-quality teacher rationales, the
augmented model achieves performance comparable to the query-only Stage-II
model on SkillReason-Bench and SkillRet. On SkillReason-Bench, the two
settings differ only marginally across all metrics, while their results on
SkillRet are also nearly identical. Teacher CoT produces a clearer improvement
on SRA-Bench, indicating that explicit reasoning can still provide additional
information for some query distributions.

Overall, the query-only 0.6B model recovers most of the benefit obtainable
from capability rationales generated by a much stronger 235B teacher. This
suggests that the Stage-I reasoning supervision and Stage-II
retrieval-guided optimization successfully internalize capability reasoning
into the query representation. As a result, SkillReason can directly infer
the capabilities implied by a user request without explicitly generating a
reasoning trace at inference, while achieving results close to those obtained
with external teacher-generated CoT.

\subsection{SkillBench Retrieval Results}
\label{app:skillbench-results}

SkillBench provides a complementary evaluation of retrieval for compound
requests. Unlike the concise and underspecified queries emphasized by
SkillReason-Bench, SkillBench requests typically describe the task context,
desired output, constraints, and multiple required capabilities in detail.
It therefore tests whether a retriever can identify a complete set of
complementary skills for a single request. Because each split contains only
75 queries, we treat these results as a diagnostic analysis and draw the
primary conclusions from the larger retrieval benchmarks.

Table~\ref{tab:skillbench-additional} reports results for the Easy and Hard
splits. Each split contains 24 single-skill and 51 multi-skill queries.
R@1 and R@10 are fractional recall averaged over all queries, while
Multi FC@10 denotes FullCoverage@10 on the multi-skill subset, measuring the
percentage of queries for which all annotated skills are retrieved within
the top 10. All methods use the released benchmark protocol, identical query
instructions, and the official relevance annotations.

\begin{table*}[!t]
  \centering
  \caption{SkillBench retrieval results (\%). R@1 and R@10 denote averaged
  fractional recall, while Multi FC@10 measures complete top-10 coverage of
  all annotated skills for multi-skill requests.}
  \label{tab:skillbench-additional}
  \scriptsize
  \setlength{\tabcolsep}{3.5pt}
  \resizebox{\textwidth}{!}{%
  \begin{tabular}{@{}lrrrrrr@{}}
    \toprule
    \textbf{Retriever}
      & \textbf{Easy R@1} & \textbf{Easy R@10}
      & \textbf{Easy Multi FC@10}
      & \textbf{Hard R@1} & \textbf{Hard R@10}
      & \textbf{Hard Multi FC@10} \\
    \midrule
    Qwen3-Embedding-0.6B & 32.50 & 61.04 & 23.53 & 30.17 & 57.42 & 17.65 \\
    Qwen3-Embedding-4B   & 33.46 & 64.15 & 31.37 & 28.57 & 59.97 & 31.37 \\
    Qwen3-Embedding-8B   & 34.21 & 67.63 & 33.33 & 28.21 & 63.96 & 27.45 \\
    SkillRet-Embedding-0.6B & 35.17 & 68.73 & 31.37 & 31.95 & 65.68 & 23.53 \\
    SkillRet-Embedding-8B   & 28.72 & 62.98 & 31.37 & 28.17 & 59.34 & 21.57 \\
    SkillRouter-Embedding-0.6B & \textbf{37.09} & 67.96 & 31.37
      & \textbf{34.87} & \textbf{66.07} & 29.41 \\
    SkillReason-0.6B & 34.72 & 67.95 & 37.25
      & 31.06 & 62.60 & 29.41 \\
    SkillReason-4B & 31.77 & \textbf{70.02} & \textbf{41.18}
      & 27.28 & 65.08 & \textbf{35.29} \\
    \bottomrule
  \end{tabular}%
  }
\end{table*}

The results reveal a clear distinction between top-rank retrieval and
complete capability coverage. SkillRouter-Embedding-0.6B achieves the
strongest R@1 on both splits, indicating that it is more likely to place one
annotated skill at the top of the ranking. SkillReason, however, performs
particularly well when success requires retrieving the complete set of skills
needed by a compound request. SkillReason-4B obtains the best Multi FC@10 on
both Easy and Hard and also achieves the best Easy R@10. On the Hard split,
its R@10 remains competitive with the strongest baselines while retaining the
highest complete-coverage score.

The same pattern is visible at the 0.6B scale. SkillReason-0.6B achieves the
highest Easy Multi FC@10 among the 0.6B retrievers and ties SkillRouter on
Hard Multi FC@10, despite not obtaining the best R@1. Thus, the improvement
is concentrated in recovering multiple complementary skills within the
retrieval set rather than in prioritizing a single annotated skill at rank
one. This result is consistent with the purpose of capability reasoning:
instead of matching a request to one dominant surface-level capability, the
model is encouraged to identify the collection of capabilities required to
complete the task.

SkillBench differs from SkillReason-Bench because its requests often state
the relevant context and sub-requirements explicitly. The gains in full
coverage therefore suggest that the learned capability reasoning transfers
beyond underspecified requests and also benefits explicit but compositional
multi-skill retrieval. Given the small number of queries in each split, we
view this result as supporting evidence rather than a primary benchmark
conclusion.

\subsection{Online Retrieval and Reranking Latency}
\label{app:efficiency}

We measure online latency on one NVIDIA A800-SXM4-80GB GPU using 100 fixed
SkillRet queries (seed 20260720), the official query instruction, batch size
one, and a maximum input length of 4096. We perform ten warm-up queries for
retrievers and two for rerankers, synchronize CUDA around every timed region,
and report statistics over individual queries. Skill documents are encoded
once and cached; model loading, corpus encoding, and index construction are
therefore excluded. Online retrieval includes query encoding and exact
inner-product top-20 search over all 6{,}660 normalized corpus embeddings. A
retrieve--rerank pipeline additionally scores all 20 candidates in one
candidate batch. All measurements use PyTorch 2.6.0 with CUDA 11.8 and run
without concurrent GPU workloads.

\begin{table*}[!tbp]
  \centering
  \caption{Batch-1 online latency on SkillRet (ms). Retrieval includes query
  encoding and exact top-20 search; for Self-CoT augmentation, it includes
  augmented-query encoding and search. Generation denotes capability-rationale
  generation, Rank denotes top-20 reranking, and Total is the mean latency of
  the complete inference path.}
  \label{tab:online-efficiency}
  \small
  \setlength{\tabcolsep}{7pt}
  \begin{tabular}{@{}llrrrr@{}}
    \toprule
    \textbf{Path} & \textbf{Model(s)} & \textbf{Retrieval}
      & \textbf{Generation} & \textbf{Rank} & \textbf{Total} \\
    \midrule
    Retrieval only & Qwen3-Embedding-0.6B
      & 31.69 & -- & -- & 31.69 \\
    & SkillReason-Emb-0.6B
      & 31.58 & -- & -- & 31.58 \\
    & Qwen3-Embedding-4B
      & 144.41 & -- & -- & 144.41 \\
    & SkillReason-Emb-4B
      & 143.93 & -- & -- & 143.93 \\
    \midrule
    Retrieve--rerank & Qwen3-Emb-0.6B $\times$ Qwen3-Rank-0.6B
      & 31.69 & -- & 1904.43 & 1936.12 \\
    & SkillReason-Emb-0.6B $\times$ SkillReason-Rank-0.6B
      & 31.58 & -- & 1911.84 & 1943.42 \\
    & Qwen3-Emb-4B $\times$ Qwen3-Rank-4B
      & 144.41 & -- & 5176.41 & 5320.82 \\
    & SkillReason-Emb-4B $\times$ SkillReason-Rank-4B
      & 143.93 & -- & 4710.06 & 4853.99 \\
    \midrule
    Query-only & SkillReason-0.6B
      & 32.27 & -- & -- & 32.27 \\
    Self-CoT augmentation & SkillReason-0.6B
      & 49.79 & 3101.31 & -- & 3151.10 \\
    Query-only & SkillReason-4B
      & 144.65 & -- & -- & 144.65 \\
    Self-CoT augmentation & SkillReason-4B
      & 217.93 & 4019.34 & -- & 4237.27 \\
    \bottomrule
  \end{tabular}
\end{table*}

Table~\ref{tab:online-efficiency} shows that reasoning-oriented training
introduces no measurable online retrieval overhead. SkillReason and the
corresponding Qwen3-Embedding backbones exhibit nearly identical latency at
both model scales, while exact vector search itself requires only
0.12--0.17 ms. Online retrieval efficiency is therefore determined primarily
by model scale rather than by the SkillReason training objectives.

Explicit CoT generation, in contrast, introduces substantial online cost.
For SkillReason-0.6B, Self-CoT augmentation increases the mean end-to-end
latency from 32.27 ms to 3151.10 ms, while for SkillReason-4B it increases
from 144.65 ms to 4237.27 ms. Rationale generation dominates the augmented
inference path, increasing latency by approximately one to two orders of
magnitude, whereas encoding the longer augmented query contributes only a
small fraction of the additional cost.

Reranking also introduces considerable latency, but provides substantially
larger retrieval improvements than inference-time CoT augmentation. It
therefore offers a more effective use of additional online computation.
Together with the limited gains from Self-CoT reported above, these results
highlight the practical advantage of internalizing capability reasoning
during training: SkillReason retains backbone-level retrieval latency while
avoiding expensive online rationale generation.
\section{SkillReason-Bench Construction Details}
\label{app:benchmark}

This section provides additional details on the construction and quality
control of SkillReason-Bench. The benchmark contains 3{,}729 queries and a
shared retrieval corpus of 61{,}228 skills spanning nine domains. Each query
is associated with one annotated target skill and describes the desired
outcome without revealing the skill name or complete execution procedure.
The final English requests contain 33.2 words on average. The core benchmark
construction instructions and output schemas are provided in
Section~\ref{app:benchmark-construction-prompts}.

\subsection{Construction Overview}
\label{app:benchmark-construction-protocol}

The construction pipeline consists of four stages: skill collection and
normalization, target-skill screening, implicit-query generation, and
multi-stage quality control. LLMs are used for target screening, query
generation, query--target validation, and query-specific alternative-skill
assessment. Deterministic operations, including corpus cleaning, identifier
validation, lexical leakage checks, and exact deduplication, are implemented
programmatically.

DeepSeek-R1 and Claude Haiku 4.5 independently review the final candidates
using a rubric covering naturalness, informativeness, skill necessity,
implicitness, and target consistency; disagreements are adjudicated by human
reviewers. The review retains natural requests that require specialized
capabilities and have a well-supported and stable target annotation, while
removing generic requests, explicit target or procedure leakage, weak
query--target associations, and requests with excessive ambiguity among
plausible targets.

\subsection{Skill Collection and Normalization}
\label{app:skill-collection}

We collect publicly available agent skills from GitHub repositories indexed by
SkillsMP. For each record, we retain a stable skill identifier, name,
description, full document, and repository metadata. After merging the
collected sources and deduplicating by stable identifier, the raw corpus
contains 69{,}138 unique skills.

We then remove records that cannot serve as reliable retrieval documents,
including entries with missing identifiers or names, insufficient content,
placeholder or template text, malformed fields, and unreadable documents.
This deterministic cleaning removes 7{,}910 records and produces the final
retrieval pool of 61{,}228 skills, as summarized in
Table~\ref{tab:skill-filtering}.

We retain valid skills that provide functionally related or independently
maintained implementations of overlapping workflows. Such overlap naturally
occurs in open skill repositories and preserves the difficulty of retrieving
the most appropriate capability from a large and partially redundant corpus.
Semantic similarity alone is therefore not used as a reason to remove a skill
from the retrieval pool.

\begin{center}
  \captionof{table}{Skill-corpus filtering statistics.}
  \label{tab:skill-filtering}
  \small
  \begin{tabular}{@{}lr@{}}
    \toprule
    \textbf{Stage} & \textbf{Count} \\
    \midrule
    Collected unique skills & 69{,}138 \\
    Removed by deterministic corpus cleaning & 7{,}910 \\
    Final retrieval pool & 61{,}228 \\
    \bottomrule
  \end{tabular}
\end{center}

\subsection{Target Screening and Implicit Query Generation}
\label{app:query-generation}

Before query generation, we use DeepSeek-R1 to screen the cleaned pool for
target skills with concrete, specialized, and executable capabilities. Of the
61{,}228 entries in the retrieval pool, 60{,}928 satisfy the minimum content
requirements for target screening, and 55{,}483 are retained as potential
target skills. We exclude generic assistant behaviors, template-only
documents, skills whose usefulness depends on unavailable private context,
and skills for which a realistic request would reduce to a direct keyword
match.

For each retained target, an LLM receives the skill name, description, and
full document and produces one concise user request. The generation instruction
asks for one to three natural sentences describing the user's goal, available
inputs, constraints, or desired deliverable. It prohibits mentioning the
target skill, repository, package name, or the word ``skill'', as well as
copying distinctive trigger phrases or enumerating the complete execution
procedure. Domain terminology is permitted when it would naturally appear in
a real user request.

The generator also produces a short capability-level rationale describing the
capabilities implied by the request. This rationale is used during construction
and auditing but is not released as an additional relevance label. Query
generation produces 51{,}164 parseable records before subsequent quality
filtering.

\subsection{Quality Control and Human Adjudication}
\label{app:quality-control}

The quality-control pipeline addresses three distinct sources of noise:
duplicate user requests, weak query--target associations, and ambiguity
introduced by alternative skills in the retrieval pool. These checks operate
on the query annotations and do not remove valid documents from the shared
skill corpus.

\paragraph{Deterministic filtering and query deduplication.}
After removing records with missing fields, invalid target identifiers,
abnormal lengths, parsing failures, and normalized exact duplicates, 43{,}885
candidates remain. A subsequent lexical leakage check removes 73 requests
that directly contain the target skill name, leaving 43{,}812 records.

Near-duplicate requests are identified using lexical overlap, BM25 retrieval,
and dense query neighbors. A pair is treated as a duplicate only when the
signals jointly indicate that the requests describe the same scenario,
inputs, constraints, and deliverable. Shared terminology or membership in the
same domain is not sufficient evidence for removal, and this process does not
modify the retrieval pool.

\paragraph{Query--target validation.}
We next evaluate whether the annotated target provides the capabilities
required by the request rather than merely sharing related terminology. A
retained request must require a specialized capability, toolchain, or domain
workflow; specify a concrete outcome, context, constraint, or deliverable;
avoid revealing the target or its complete procedure; and have a sufficiently
clear capability boundary.

This stage rejects simple writing or formatting requests, broad advice,
explicit target references, weak query--target associations, and requests
with excessive ambiguity among plausible targets. The strict quality screen
retains 36{,}512 of the 43{,}812 candidates. The largest rejection categories
include simple LLM-solvable tasks (4{,}297), vague or generic requests
(1{,}435), explicit keyword matches (779), and requests with too many
plausible target skills (677).

\paragraph{Alternative-skill audit.}
Because the retrieval pool contains overlapping implementations, we also
audit whether other skills could satisfy a particular request. Candidate
alternatives are proposed using normalized names, document hashes, repository
metadata, name and document overlap, text containment, rare-token matches,
and dense neighbors. These signals are used only to identify candidates for
review and do not directly determine relevance.

For each query--candidate pair, an LLM compares the request, annotated target,
and candidate skill and assigns one of three labels:
\texttt{positive}, \texttt{ignore}, or \texttt{hard\_negative}.
A \texttt{positive} candidate can independently satisfy the core request;
\texttt{ignore} denotes a related or potentially useful skill that does not
fully cover the request; and \texttt{hard\_negative} denotes a superficially
similar skill with an incompatible capability boundary. The decision is based
on substitutability for the specific request rather than general topical
similarity.

In an intermediate audit of 5{,}320 sampled queries, 1{,}380 queries produced
potentially ambiguous dense neighbors, resulting in 24{,}697 reviewed
query--candidate pairs. Within this audit set, 10{,}077 candidate pairs are
assigned as \texttt{positive}, 4{,}941 as \texttt{ignore}, and 9{,}679 as
\texttt{hard\_negative}. These labels are used only to identify annotation
ambiguity and calibrate the subsequent filtering process; they are not added
as released benchmark labels. Dense similarity alone is never used as a
relevance label.

\paragraph{Balanced sampling and final review.}
After the preceding filters, we perform domain-balanced sampling to obtain
6{,}000 candidates and subsequently narrow the set to 4{,}000. Twenty-eight
requests with unstable target assignments are replaced by same-domain
requests with clearer query--target relationships.

DeepSeek-R1 and Claude Haiku 4.5 independently review the 4{,}000 candidates
using a rubric covering naturalness, informativeness, skill necessity,
implicitness, and target consistency. Human reviewers adjudicate disagreements
between the two models.
The final review removes 271 records, including 112 simple tasks, 58
multi-answer requests, 42 target mismatches, 21 target-scope risks, and 17
explicit matches. This process yields the final set of 3{,}729 benchmark
queries.

\subsection{Released Format and Query Characteristics}
\label{app:benchmark-format}

Each released record contains a user request, one annotated target-skill
identifier, and a reference to the shared retrieval corpus. Thus, each query
is associated with one stable target annotation, while functionally related
skills may remain in the shared retrieval pool. The sensitivity of the
evaluation to unannotated but valid alternatives is examined separately in
Section~\ref{app:alternative-gold}.

The benchmark queries are designed around a gap between the outcome stated by
the user and the procedural capabilities encoded in the target skill. For
example, a request to transform a local tutorial video into timestamped
chapter notes does not explicitly mention speech transcription, temporal
alignment, or chapter segmentation, although these capabilities are needed to
identify the appropriate skill. Similar capability gaps occur across software
maintenance, engineering diagnosis, biomedical analysis, legal document
comparison, and the other covered domains.

\section{Training Data Construction}
\label{app:training}

\subsection{Stage-I and Stage-II Data}
\label{app:training-data}

We construct the training data from the public corpus of 65{,}714 skills
released by SkillRouter, which is sourced from the
Claude Skill Registry. The training and evaluation queries are constructed
through separate pipelines and collected from different sources. No query
from SkillReason-Bench, SRA-Bench, or SkillRet is used as a training source
in either stage. The core training-data prompts and output schemas are
provided in Section~\ref{app:training-data-prompts}.

\paragraph{Query synthesis.}
A skill is eligible as a training target only when its stable identifier
occurs in both the released corpus and the corresponding embedding index.
Claude Sonnet 4.6, with thinking enabled, receives the eligible skill's name,
description, and document and generates three candidate user requests per
skill. Each request must describe a concrete functional need rather than name,
summarize, or directly restate the skill. This process covers 30{,}855 target
skills and produces 92{,}565 single-target query--skill pairs. Subsequent
quality filtering and coverage-constrained sampling retain at most two
requests for each target skill. Stable identifiers, rather than display names,
are used for all subsequent exclusion, deduplication, and sampling operations.

\paragraph{Potential false-negative graph construction.}
Open skill repositories contain forks, renamed copies, and independently
maintained implementations of overlapping functionality. Treating such skills
as negatives can introduce contradictory supervision. We therefore construct
a graph of skill pairs that should be excluded from negative sampling.

We first $\ell_2$-normalize Qwen3-Embedding-0.6B representations of all
training skills and use an exact FAISS inner-product index to retrieve the
top 50 neighbors of each skill. After removing self-pairs and duplicate
undirected pairs, pairs with cosine similarity greater than 0.85 are retained
as candidates. For each candidate, we additionally compute character-trigram
Jaccard similarity between the two skill documents.
A score greater than 0.60 identifies a high-confidence content copy. Pairs
below this threshold are retained only as conservative similarity-based
exclusion edges because their dense similarity remains high; they are not
treated as relevance labels.

The resulting graph contains 982 high-confidence copy pairs and 1{,}007
similarity-based exclusion pairs, for a total of 1{,}989 undirected edges
involving 3{,}300 skills. We store the graph as a symmetric adjacency map.
All documents remain in the training corpus and may serve as positives for
their own queries, but a skill connected to the current positive is excluded
from every negative source. This preserves realistic corpus redundancy while
avoiding potentially contradictory negative supervision.

\paragraph{Four-source negative composition.}
Each generated request is paired with a fixed, ordered set of ten explicit
negatives. The quotas and purposes are shown in Table~\ref{tab:negative-composition}.
The first two sources are retrieval-guided hard negatives: one is close to the
positive document, and the other is close to the request. The category samples
add controlled within-domain confusions, while the final cross-category sample
provides a comparatively easy separation signal. The order is deterministic
within a record (skill neighbors, query neighbors, same-category samples,
then the cross-category sample), which makes the composition auditable and
keeps the raw and rationale-augmented views paired with exactly the same
candidate set.

\begin{table}[!t]
  \centering
  \caption{Composition of the ten explicit negatives per training
  query. ``Occurrences'' counts negative positions over the 30{,}000 selected
  rows; it is not a count of unique skill IDs.}
  \label{tab:negative-composition}
  \scriptsize
  \setlength{\tabcolsep}{3pt}
  \resizebox{\columnwidth}{!}{%
  \begin{tabular}{@{}p{0.27\columnwidth}rrp{0.43\columnwidth}@{}}
    \toprule
    \textbf{Source} & \textbf{Per query} & \textbf{Occurrences}
      & \textbf{Construction and purpose} \\
    \midrule
    Positive-skill embedding neighbors
      & 4 & 120{,}000
      & Top admissible entries in the positive skill's embedding top-50;
        separates closely related documents. \\
    Query-retrieved neighbors
      & 3 & 90{,}000
      & Top admissible skills returned by the encoded request; approximates
        errors likely to occur at inference. \\
    Same-category samples
      & 2 & 60{,}000
      & Uniform samples from the positive skill's category prefix; teaches
        fine-grained within-domain boundaries. \\
    Cross-category sample
      & 1 & 30{,}000
      & Uniform sample from a different category prefix; supplies a stable
        broad-domain contrast. \\
    \bottomrule
  \end{tabular}
  }
\end{table}

For the first source, we L2-normalize the skill embeddings and search an
exact FAISS inner-product index. For the second source, we encode the request
with the same Qwen3-Embedding-0.6B encoder and search the full skill index;
the highest-ranked admissible results are retained. Same-category and
cross-category candidates are sampled with seed 42. The primary data therefore
use dense query retrieval rather than BM25; a BM25 replacement of only the
three query-neighbor positions was used as a controlled variant, not in the
reported Stage-I checkpoint.

\paragraph{Candidate eligibility and false-negative exclusion.}
Negative selection is sequential. Before selecting from any source, we
construct
the forbidden set
\[
  F(q)=\{s_q^+\}\cup\mathcal{N}_{\mathrm{excl}}(s_q^+),
\]
where $s_q^+$ is the labeled positive and
$\mathcal{N}_{\mathrm{excl}}(s_q^+)$ denotes the skills connected to it by
the embedding- and text-based exclusion graph. After each source is selected,
its IDs are added to the forbidden set. A candidate is accepted only if its
stable ID occurs in the corpus and index, is not in $F(q)$, and has not already
been selected for the same query. This applies identically to all four sources,
so a document cannot be counted twice under different negative types. If a
source cannot supply its quota after these exclusions, the row is not repaired
by silently reusing an ineligible document; it is removed by the subsequent
exact-count validation.

The exclusion graph is deliberately conservative. A cosine similarity above
0.85 is sufficient to block a document from negative sampling, while a
character-trigram Jaccard score above 0.60 additionally identifies a
high-confidence content copy. The graph is used only as a negative-sampling
filter and does not provide relevance annotations. Skills connected by the
graph remain in the searchable corpus and may serve as positives for their own
requests. This prevents a fork, mirror, or renamed implementation from becoming
a contradictory negative without collapsing realistic corpus redundancy.

\paragraph{Negative-set cleaning and structural validation.}
After mining, we apply deterministic row-level checks before quality selection.
The query, positive text, positive ID, negative text list, and negative ID list
must all be present. Both negative lists must contain exactly ten entries, the
IDs must be unique, the positive ID must be absent from the negative IDs, and
every ID must resolve to a non-empty corpus document. We also verify that each
serialized negative text is the canonical
\texttt{name | description[:300] | body[:2500]} representation of its ID.
Malformed JSON, missing corpus entries, empty text, duplicate IDs, positive--
negative overlap, and any residual exclusion-graph edge are rejected rather
than repaired by relabeling.

The query itself must contain 8--110 whitespace-delimited words and at most
700 characters. In the 92{,}565-row Sonnet pool, this pass removes 38 short
requests, 54 overlong requests, and 27 rows with malformed negative lists,
leaving 92{,}446 structurally valid rows. An independent audit of the final
30{,}000-row subset finds 300{,}000 negative positions with zero positive--
negative overlaps, duplicate IDs, missing corpus IDs, exclusion-graph
violations, or ID--text mismatches.

\paragraph{Quality filtering and embedding-based deduplication.}
We then score each valid row using query length, synthetic meta-language, and
token overlap with the positive and negative names/descriptions. Phrases such
as ``sample query'', ``trigger this skill'', and ``use this skill'' are
penalized. If a negative has higher lexical overlap than the labeled positive,
the row is down-ranked as a possible label-risk case, but it is not automatically
deleted: such candidates are often precisely the fine-grained hard negatives
needed for training. The deterministic score is used only to prioritize
candidates for subsequent quality review and subset selection; it is not used
as a training weight.

To reduce repeated synthetic formulations, we next encode the structurally
valid requests with the dense embedding encoder used by the retrieval-data
pipeline, apply $\ell_2$ normalization, and retrieve nearest-neighbor requests
by exact inner-product search. High-similarity candidates are checked after
text normalization so that common domain vocabulary alone does not collapse
distinct tasks. When multiple requests express the same intent and retain the
same target association, the higher-scoring and more informative request is
kept; ties are resolved by stable input order. Similar requests attached to
materially different target capabilities are instead preserved for
query--target review. This step is separate from the query embeddings used to
mine the three query-retrieved negative positions in
Table~\ref{tab:negative-composition}.

DeepSeek-R1 then reviews the remaining requests and removes outputs that are
invalid, redundant, unnatural, insufficiently informative, or inconsistent
with their target skills. Thus, the three requests initially generated for
each skill provide alternatives for quality selection rather than three
guaranteed training instances.

\paragraph{Coverage-constrained selection.}
After quality filtering, the remaining candidates are grouped by the category
prefix of the positive skill identifier. Smaller categories first reserve up
to 500 high-scoring records, with at most two requests retained for each target
skill. A deterministic global pass then fills the training set to 30{,}000
records, with no category contributing more than 40\% of the final set.

The resulting Stage-I data cover 17{,}427 positive skills. Among them, 4{,}854
appear once and 12{,}573 appear twice. The selected requests contain 47.55
words on average. At training time, the separated batch sampler prioritizes
batches without repeated groups or positive--negative collisions, so explicit
and in-batch negatives do not create avoidable contradictions.

\paragraph{Capability-level rationale annotation.}
For each of the 30{,}000 selected requests, Claude Sonnet 4.6 generates a
concise capability-level rationale. In our implementation, this rationale is
the teacher reasoning trace $c^*$: it summarizes the capabilities required by
the request rather than providing a step-by-step task solution. The model
receives the request and its target skill but does not observe the negative
skills. It does not name a skill or directly solve the task. The rationale
provides an auxiliary view of the same query--skill pair and is used for
reasoning-augmented contrastive learning, retrieval distribution alignment,
and teacher-forced generation in Stage I.

\paragraph{Stage-II implicit-query data.}
Claude Haiku 4.5 rewrites each Stage-I request into a more natural and implicit
form while preserving its task goal, available inputs, constraints, desired
deliverable, target skill, and ten negative skills. The rewriting suppresses
unnecessary references to skills, tools, algorithms, APIs, packages, and
procedural expressions copied from the skill document.

Structured parsing retains 29{,}995 of the 30{,}000 rewritten requests, with
five records removed because of parsing failures. Each Stage-II record stores
the original request, implicit rewrite, teacher rationale, target skill, and
unchanged negative set. Keeping the labels and candidates fixed ensures that
the retrieval rewards compare sampled reasoning traces under the same
retrieval setting.

\subsection{Training--Test Query Isolation}
\label{app:training-test-isolation}

We audit the combined set of 30{,}000 Stage-I queries and 29{,}995 Stage-II
implicit queries against every evaluation query in SkillReason-Bench,
SRA-Bench, and SkillRet. For exact matching, we apply Unicode NFKC
normalization, case folding, punctuation removal, and whitespace
normalization. To identify lightly rewritten requests, we represent each
normalized query using character 3- to 5-grams and retrieve its nearest
training query by cosine similarity. We use 0.85 as the threshold for
near-duplicate review.

\begin{center}
  \captionof{table}{Training--test query-isolation audit. Comparisons use the
  combined set of 59{,}995 Stage-I and Stage-II training queries.}
  \label{tab:training-test-isolation}
  \small
  \resizebox{0.98\columnwidth}{!}{%
  \begin{tabular}{@{}lrrrr@{}}
    \toprule
    \textbf{Benchmark} & \textbf{Test Queries}
      & \textbf{Exact Matches} & \textbf{$\geq 0.85$}
      & \textbf{Max. Similarity} \\
    \midrule
    SkillReason-Bench & 3{,}729 & 0 & 0 & 0.6431 \\
    SRA-Bench         & 5{,}400 & 0 & 0 & 0.6643 \\
    SkillRet          & 4{,}997 & 0 & 0 & 0.7409 \\
    \bottomrule
  \end{tabular}%
  }
\end{center}

No evaluation query exactly matches a training query, and none reaches the
0.85 near-duplicate threshold. As a stricter diagnostic, we lower the review
threshold to 0.70 and manually inspect all nine unique candidate pairs, all
of which occur in SkillRet. These pairs describe related capabilities but
differ in their concrete inputs, scope, constraints, or desired deliverables.
None is a copied or lightly rewritten training request.

\section{Implementation Details}
\label{app:implementation}

\subsection{Input Formatting}
\label{app:input-format}

Skill documents are serialized as
\texttt{name | description[:300] | body[:2500]}. A Stage-I record stores the
raw request, positive text and ID, ten negative texts and IDs, and its
capability rationale. The raw retrieval view prepends the Qwen3 instruction
``Instruct: Given a task description, retrieve the most relevant skill
document that would help an agent complete the task'' followed by ``Query:''
to the request. The privileged retrieval view keeps the same instruction,
inserts the rationale
under ``Query analysis:'', and ends with the summary anchor ``In summary, the
retrieval representation for this task is:''. The teacher view used for
distribution alignment uses the corresponding rationale-aware retrieval suffix
``Based on the user task and the analysis above, retrieve the most relevant
skill document that would help the agent complete the task.'' These concrete
templates instantiate the abstract $[q;c^*]$ notation in the main paper. The
raw view is used for $\mathcal{L}_{\mathrm{raw}}^{\mathrm{CL}}$; the
rationale-augmented retrieval view is used for
$\mathcal{L}_{\mathrm{cot}}^{\mathrm{CL}}$; and the teacher view is used to
construct the privileged ranking distribution in $\mathcal{L}_{\mathrm{KL}}$.
The two augmented views contain the same request and teacher rationale and
differ only in their retrieval-instruction suffix. The rationale-generation
prompt is given in Section~\ref{app:training-data-prompts}.
Both retrieval views use a maximum sequence length of 2{,}048 during Stage-I
training. The rationale language-modeling branch uses a maximum length of 512
and limits the generated rationale to 96 tokens. All benchmark evaluations use
the 4{,}096-token maximum context length specified by the evaluation protocol.

A Stage-II record uses the implicit rewrite as \texttt{query} and retains the
source request as \texttt{original\_query}. It also carries
\texttt{query\_capability}, the same positive and negative documents, and the
sample weight. The GRPO rollout receives only the implicit query; the original
request and teacher rationale are retained for bookkeeping and controlled
auxiliary-loss experiments rather than exposed to the reward-generation path.

\subsection{Stage-I Configuration}
\label{app:stage1-config}

Both model sizes are full-parameter fine-tuned for one epoch. We take the final
non-padding hidden state, apply L2 normalization, and use cosine similarity
without an additional projection head. The same backbone and LM head support
retrieval and capability-rationale generation. The KL weight is activated
after 30 updates.

\begin{center}
  \captionof{table}{Stage-I training configuration.}
  \label{tab:stage1-config}
  \small
  \resizebox{0.98\columnwidth}{!}{%
  \begin{tabular}{@{}lcc@{}}
    \toprule
    \textbf{Hyperparameter} & \textbf{0.6B} & \textbf{4B} \\
    \midrule
    Initialization & Qwen3-Emb-0.6B & Qwen3-Emb-4B \\
    Training records & 30{,}000 & 30{,}000 \\
    Effective query batch & 256 & 256 \\
    Epochs / optimizer updates & 1 / 117 & 1 / 117 \\
    Optimizer / learning rate & AdamW / $2\times10^{-5}$ & Adafactor / $1\times10^{-5}$ \\
    Maximum retrieval length & 2{,}048 & 2{,}048 \\
    Contrastive temperature $\tau$ & 0.05 & 0.05 \\
    $(\lambda_{\rm raw},\lambda_{\rm cot})$ & $(1,1)$ & $(1,1)$ \\
    $(\lambda_{\rm KL},T)$ & $(5,1)$ & $(5,1)$ \\
    $\lambda_{\rm LM}$ & 0.2 & 0.2 \\
    \bottomrule
  \end{tabular}%
  }
\end{center}

We train in BF16 on eight GPUs and use gradient caching to realize the listed
effective batch. The final save contains both a
SentenceTransformer-compatible encoder and a full CausalLM checkpoint; Stage
II is initialized from the latter.

\subsection{Stage-II Configuration}
\label{app:stage2-config}

The actor is initialized from the matching Stage-I full CausalLM. A frozen
copy of the matching Stage-I encoder serves as the reward retriever and is
used only to compute retrieval margins. A separate frozen copy of the
Stage-I full CausalLM is used as the GRPO reference policy for reference-log
probabilities and KL regularization. Thus, the trainable actor, frozen reward
retriever, and frozen reference policy are three distinct roles. The primary
checkpoint uses the clipped GRPO implementation adopted in our experiments,
with auxiliary SFT losses disabled.

\begin{center}
  \captionof{table}{Stage-II retrieval-guided GRPO configuration.}
  \label{tab:stage2-config}
  \small
  \resizebox{0.98\columnwidth}{!}{%
  \begin{tabular}{@{}lcc@{}}
    \toprule
    \textbf{Hyperparameter} & \textbf{0.6B} & \textbf{4B} \\
    \midrule
    Training records & 29{,}995 & 29{,}995 \\
    Epochs / updates & 1 / 267 & 1 / 267 \\
    Query-group batch & 112 & 112 \\
    Rollouts per query $G$ & 8 & 8 \\
    Optimizer / learning rate & AdamW / $2\times10^{-6}$ & AdamW / $2\times10^{-6}$ \\
    Prompt / response limit & 1{,}024 / 192 & 1{,}024 / 192 \\
    Temperature / top-$p$ & 0.8 / 0.95 & 0.8 / 0.95 \\
    Policy clip / dual clip & 0.2 / 3.0 & 0.2 / 3.0 \\
    Reference KL coefficient & 0.001 & 0.001 \\
    \bottomrule
  \end{tabular}%
  }
\end{center}

The Stage-I sampler and the Stage-II data loader drop incomplete final query
batches; Stage-I also drops an incomplete gradient-accumulation window. This
accounts for the floor-based update counts of 117 and 267 in the two
retriever stages.

The reward uses
$\lambda_{\mathrm{margin}}=\lambda_{\mathrm{gain}}=1$ and
$\tau_{\mathrm{margin}}=\tau_{\mathrm{gain}}=0.05$, with no discrete gain
bonus. The length term is
\begin{equation}
P_{\mathrm{len}}(c)=
\min\!\left(0.4,\,0.01\max(0,|c|-128)\right),
\label{eq:app-length-penalty}
\end{equation}
where $|c|$ counts response tokens only. We use group-relative advantage
normalization over the eight rollouts for each query. In the clipped GRPO
implementation, the normalized advantage for rollout $g$ is
\begin{equation}
  A^{(g)}=
  \frac{R^{(g)}-\operatorname{mean}_{j}R^{(j)}}
  {\operatorname{std}_{j}R^{(j)}+\varepsilon},
  \qquad \varepsilon=10^{-6},
  \label{eq:app-grpo-advantage}
\end{equation}
where the mean and standard deviation are computed within the query group.
We use BF16 FSDP for actor and reference-policy computation and synchronous
vLLM rollouts. All runs use one node with eight NVIDIA A800 80GB GPUs.

\subsection{Stage-II Reward Candidate Pool}
\label{app:negative-sampling}

Each implicit rewrite inherits the positive and ten explicit negatives of its
Stage-I source record. In the implementation, the negative set
$\mathcal{N}_q$ in the main paper is instantiated from a batch-level candidate
pool. For a batch of 112 unique queries, the frozen reward encoder embeds the
concatenated candidate groups once, producing at most
$112\times11=1{,}232$ document occurrences. For a query representation $x$,
the reward margin is computed as
\begin{equation}
  \mathcal{N}_q=\mathcal{C}_{\mathrm{batch}}\setminus\{s_q^+\},
  \qquad
  m(x)=h_\phi(x,s_q^+)-\max_{s\in\mathcal{N}_q}h_\phi(x,s),
  \label{eq:app-batch-hard-negative}
\end{equation}
where $\mathcal{C}_{\mathrm{batch}}$ is the concatenation of all candidate
groups and the current query's labeled positive occurrence is masked. Thus the
hard negative is the highest-scoring document among its explicit negatives and
the documents contributed by the other query groups. Each record retains its
fixed ten explicit negatives, while the other groups supply the in-batch
candidate pool used for the margin computation. The raw query is encoded once
to obtain the baseline margin, whereas each of its eight sampled rationales
forms a different reasoning-augmented query. All rollouts consequently share
the same labels, documents, and raw-query baseline; only the sampled rationale
changes.

\subsection{Reranker Training}
\label{app:reranker-training}

\paragraph{Multi-skill supervision.}
Stage I and Stage II use single-target supervision, but reranking must also
order several complementary skills for compound agent tasks. We therefore
construct an additional multi-skill set from 27{,}583 grounded synthetic agent
tasks annotated with two to five target skills. Structural checks require a
valid query, distinct target IDs present in the corpus, and at least one target
skill retrievable by the seed retriever. We score the remaining 24{,}583
candidates using query specificity, target coverage under a seed retriever,
and lexical alignment, then apply category and per-skill caps to select
20{,}000 query groups.

For each selected query, we initially mine eight query-retrieved negatives,
one same-category negative, and one cross-category negative. All target skills,
their false-negative neighbors, and candidates whose cosine similarity to any
target exceeds 0.85 are excluded. An independent LLM-based item-level quality
audit using a fixed rubric subsequently labels every query--skill relation against
the query and the candidate's skill name, description, and 700-character body
excerpt. The fixed audit rubric
(Section~\ref{app:reranker-positive-audit-prompt})
assigns one of the exact labels \texttt{strong\_positive},
\texttt{auxiliary\_positive}, \texttt{borderline\_positive},
\texttt{weak\_generic}, or \texttt{wrong\_positive}. A
\texttt{strong\_positive} directly satisfies a core capability required by
the task; an \texttt{auxiliary\_positive} is genuinely useful for a secondary
subtask but is not independently required to complete the request. Removing
weak and wrong relations leaves 19{,}985 groups with 53{,}175 retained
positive relations. For the primary reranker, we use only
\texttt{strong\_positive} relations, yielding 19{,}903 multi-skill groups and
51{,}564 positive relations. Combining these groups with the 30{,}000 Stage-I
single-skill queries gives 49{,}903 distinct reranker-training queries.

\paragraph{Retriever-conditioned candidate groups.}
The original ten negatives are not reused directly by the reranker. Instead,
the corresponding 0.6B or 4B Stage-II retriever encodes all 49{,}903 queries
and searches the full 65{,}714-skill corpus. We retain its actual top-20 list
and label every retrieved member of the audited target set as positive;
unretrieved targets are not inserted artificially. Groups in which no target
is retrieved are discarded, so the reranker is trained on the candidate
distribution it will encounter at inference time.

We then remove negative candidates that duplicate or closely overlap a
positive. The filters combine normalized-name Jaccard at 0.60, Jaccard
similarity over lowercased alphanumeric skill-ID components at 0.80,
document-token Jaccard at 0.60, containment checks, and embedding cosine
similarity at 0.85. A retained group must contain at least eight negatives.
Because the two Stage-II retrievers produce different candidate lists, the
final sets contain 46{,}912 groups for 0.6B and 47{,}628 for 4B. Before this
candidate cleaning, 94.34\% of the 0.6B groups and 95.83\% of the 4B groups
have at least one audited positive in their retrieved top-20 list.

\paragraph{Listwise training setup.}
For each query--skill pair, the causal-LM cross-encoder receives the Qwen3
reranking instruction and scores the final answer position using the relative
``yes''-versus-``no'' token logit. Let $r_\psi(q,s)$ denote this scalar score,
let $\mathcal{C}_q$ be the retrieved candidate group for query $q$, and let
$y_{q,s}>0$ denote the audited positive weight for a retained positive
candidate, with $y_{q,s}=0$ for all other candidates. We first normalize the
positive weights within each candidate group:
\begin{equation}
\bar y_{q,s}
=
\frac{y_{q,s}}
{\sum_{u\in\mathcal{C}_q} y_{q,u}}.
\label{eq:app-normalized-positive-weight}
\end{equation}

The grouped multi-positive listwise objective is then defined as
\begin{equation}
\mathcal{L}_{\mathrm{rank}}
=
-\frac{1}{|\mathcal{G}|}
\sum_{q\in\mathcal{G}}
\sum_{s\in\mathcal{C}_q}
\bar y_{q,s}
\log
\frac{\exp\!\left(r_\psi(q,s)\right)}
{\sum_{s'\in\mathcal{C}_q}
\exp\!\left(r_\psi(q,s')\right)}.
\label{eq:app-listwise-reranking}
\end{equation}

Here, $\mathcal{G}$ denotes the set of candidate groups in the minibatch that
contain at least one retained positive. The primary run combines this listwise
term with query-analysis distillation and query-rationale language modeling,
using weights of $1.0$ and $0.2$, respectively. These auxiliary objectives act
only on the query representation and do not modify the candidate labels.
Data sizes and common optimization settings are reported in
Table~\ref{tab:reranker-config}.

\begin{center}
  \captionof{table}{Primary reranker training configuration.}
  \label{tab:reranker-config}
  \small
  \resizebox{0.98\columnwidth}{!}{%
  \begin{tabular}{@{}lcc@{}}
    \toprule
    \textbf{Hyperparameter} & \textbf{0.6B} & \textbf{4B} \\
    \midrule
    Initialization & Qwen3-Rank-0.6B & Qwen3-Rank-4B \\
    Query groups & 46{,}912 & 47{,}628 \\
    Maximum candidates & 20 & 20 \\
    Effective group batch & 128 & 128 \\
    Epochs / optimizer updates & 1 / 367 & 1 / 373 \\
    Optimizer / learning rate & AdamW / $10^{-5}$ & AdamW / $10^{-5}$ \\
    Pair input limit & 4{,}096 & 4{,}096 \\
    \bottomrule
  \end{tabular}%
  }
\end{center}

For the reranker, the sampler has one group per device and accumulates 16 local
steps; its training loop still applies the final partial accumulation window.

We train the rerankers in BF16 on eight GPUs. A deterministic source-aware
sampler distributes single- and multi-skill groups across local batches while
preserving their overall proportions. At inference, rerankers receive only the
raw query and each retrieved document and do not generate or consume a
capability rationale.

\section{Benchmark and Training Data Examples}
\label{app:data-examples}

This section presents representative examples from the released English
benchmark and the finalized training data. The benchmark requests and target
identifiers, together with the Stage-I query, capability-level rationale, and
Stage-II rewrite in the training example, are reproduced verbatim. Skill
documents and the functions of the displayed negative skills are summarized
for readability. The benchmark examples are drawn from the held-out evaluation
set and are not used for training.

\subsection{Representative SkillReason-Bench Records}

The following examples illustrate the gap between the outcome stated by the
user and the specialized capability represented by the annotated target.
Both examples are released records with stable target assignments after final
review, although functionally related skills may remain in the retrieval pool.

\begin{casebox}[Benchmark Example 1: Stalled Kubernetes Resource Adjustment (UID 40482)]
\textbf{Released user request.} ``Some Pods in the cluster keep getting
terminated because of insufficient memory. Automatic resource adjustment is
clearly enabled, but the requested CPU and memory never change. How should I
investigate the cause?''

\medskip
\textbf{Annotated target.} \texttt{debug-vpa}. This skill diagnoses failures
in the Kubernetes Vertical Pod Autoscaler by checking Metrics Server
availability, VPA components and conditions, recommender logs, and updater
eviction events.

\medskip
\textbf{Implicit capability gap.} The request reports out-of-memory
terminations and unchanged resource requests despite automatic adjustment, but
does not name the Vertical Pod Autoscaler or any of its diagnostic components.
Retrieval must infer a failure in the recommendation or resource-update path
and distinguish it from generic memory tuning, horizontal autoscaling, or
Pod-failure diagnosis.
\end{casebox}

\begin{casebox}[Benchmark Example 2: Cross-Border Land-Registry Evidence (UID 10329)]
\textbf{Released user request.} ``I have a real estate power of attorney from
Spain and am preparing to use it for land registration procedures in Germany,
but I am worried the document format may not meet local requirements and need
to confirm the certification process and potential risks.''

\medskip
\textbf{Annotated target.}
\texttt{auslandsurkunden-\allowbreak apostille-\allowbreak baulast-ist}.
This skill checks the
apostille or legalization route, notarization and translation requirements,
and representation or registry evidence required when a foreign instrument is
submitted to a German land registry.

\medskip
\textbf{Implicit capability gap.} The request supplies the originating
jurisdiction, destination jurisdiction, instrument type, and intended legal
use, but does not enumerate the required checks. Retrieval must connect these
facts to cross-border document authentication and German land-registry
formalities, rather than return a generic real-estate or contract-review skill.
\end{casebox}

\subsection{Aligned Stage-I and Stage-II Training Record}
\label{app:training-data-example}

The following example shows how a selected Stage-I record is converted into
its Stage-II implicit view. The positive skill and all ten negative identifiers
are carried over unchanged from Stage I to Stage II; only the user request is
rewritten.

\begin{casebox}[Training Example: Batch-Aware RNA-seq Differential Expression]
\textbf{Stage-I query.} ``My RNA-seq experiment has samples from two tissue
types collected across three sequencing batches. I have a count matrix and a
metadata file with 'batch' and 'tissue' columns. How do I set up a multi-factor
design in Python to control for batch effects while testing for tissue-specific
differential expression, and how do I apply log fold-change shrinkage to the
results?''

\medskip
\textbf{Positive skill.} \texttt{pydeseq2}, which fits DESeq2-style
negative-binomial models to bulk RNA-seq counts in Python and supports
differential testing with batch covariates and log fold-change shrinkage.

\medskip
\textbf{Capability-level rationale.} ``The user needs skills for fitting
multi-factor negative binomial GLMs with batch correction and applying log
fold-change shrinkage to RNA-seq count data.''

\medskip
\textbf{Stage-II implicit rewrite.} ``I have RNA-seq data comparing two tissue
types that were processed in three separate sequencing runs, with a gene count
matrix and sample metadata indicating both tissue and batch. How can I identify
which genes show different expression between these tissues while accounting
for batch differences, and ensure the effect size estimates are robust against
noise? I'd prefer a Python approach.''

\medskip
\textbf{Representative mined negatives.}
\begin{itemize}
  \setlength{\itemsep}{1pt}
  \setlength{\parskip}{0pt}
  \item \texttt{bio-small-rna-seq-differential-mirna}: analyzes miRNA from
    small-RNA sequencing rather than bulk gene-expression counts.
  \item \texttt{pseudobulkdeg}: first aggregates single-cell measurements into
    pseudobulk samples, which does not match the supplied bulk RNA-seq matrix.
  \item \texttt{bio-de-results}: filters and annotates already computed
    differential-expression results but does not fit the requested model.
\end{itemize}
\end{casebox}

The rewrite replaces explicit statistical procedure names with goal-level
requirements while preserving the bulk RNA-seq modality, two tissue groups,
three sequencing batches, count matrix and metadata, batch-adjustment goal,
effect-size stabilization, and Python constraint. The three displayed
negatives are representative embedding neighbors of the positive skill in this
record. The complete ten-negative set was constructed using the four-source
composition and fixed quotas reported in
Table~\ref{tab:negative-composition}.

\section{Prompt Templates}
\label{app:prompt-templates}

For reproducibility, this section provides the core task instructions and
output schemas used for benchmark construction, data generation, and quality
assessment. Sample-specific records are replaced with brace-delimited
placeholders. Model-specific transport wrappers and few-shot demonstrations
are omitted because they do not change the evaluation criteria or required
output fields.

\subsection{Independent Quality-Audit Prompt}
\label{app:benchmark-validation-prompt}

\begin{promptbox}[System message]
You are an independent evaluator of a skill-retrieval benchmark. Evaluate each
criterion separately. Judge naturalness from the user request itself, and judge
alignment and necessity from the relationship between the request and the gold
skill. Do not treat lexical overlap, shared terminology, or polished prose
alone as evidence of quality. Do not penalize domain terminology that a real
user would naturally use. Return exactly one JSON object and no markdown.
\end{promptbox}

\begin{promptbox}[User message template]
Audit the benchmark item below.

USER REQUEST:
{query}

GOLD SKILL:
{skill_name}
{skill_description}
{skill_body}

Score each criterion from 1 (clearly fails) to 5 (clearly satisfies):

- naturalness: Judge only the user request. It should resemble a plausible
  request from a real user rather than benchmark instructions, keyword
  stuffing, or synthetic explanatory prose.

- gold_alignment: Judge whether the gold skill provides the specific
  capability, workflow, or domain procedure required by the request. Topic
  relevance or keyword overlap alone is insufficient.

- skill_necessity: Judge whether completing the request would materially
  benefit from a specialized skill, toolchain, workflow, or domain procedure.

- implicitness: Judge whether the request states its goal, context, and
  constraints without naming the target skill, repository, or a distinctive
  copied procedure.

Set leakage=true only if the request directly reveals the target skill name,
repository identifier, or a near-verbatim distinctive procedure.

Set pass=true only when naturalness>=4, gold_alignment>=4,
skill_necessity>=3, implicitness>=3, and leakage=false.

Return only JSON:
{"naturalness":1,"gold_alignment":1,"skill_necessity":1,
"implicitness":1,"leakage":false,"pass":false,
"reason":"one concise sentence"}
\end{promptbox}

\subsection{SkillReason-Bench Construction Prompts}
\label{app:benchmark-construction-prompts}

\paragraph{Anchor-skill screening.}
\begin{promptbox}[Anchor-skill screening]
You are an expert curator of an agent-skill retrieval benchmark. Given one
skill record, decide whether it can serve as an anchor for an implicit user
request.

Keep an anchor only when it describes a concrete, specialized, and executable
capability that can be inferred from a realistic user goal. Reject generic
assistant behavior, templates, placeholders, malformed or unreadable
documents, skills that require unavailable private repository context, and
skills for which a realistic request would reduce to a direct keyword match.

Set label=1 when the skill should be retained and label=0 when it should be
rejected.

Return only JSON:
{"label":1,"reason":"one concise sentence"}

SKILL RECORD:
{skill_record}
\end{promptbox}

\paragraph{Implicit-query generation.}
\begin{promptbox}[Implicit-query generation]
You create realistic benchmark queries for skill retrieval. Given a target
skill, write one concise user request that describes an outcome, input context,
constraints, or desired deliverable while leaving the required capability
implicit.

Requirements:
- Use one to three natural sentences.
- Do not mention the target skill name, repository, package, or the word
  ``skill''.
- Do not enumerate the complete toolchain or execution procedure.
- Do not copy distinctive trigger phrases from the skill document.
- Preserve domain terminology when a real user would naturally use it.
- Ensure that the request requires the target capability rather than a generic
  conversational answer.

Return only JSON:
{"query":"...","capability_reasoning":"one concise sentence"}

TARGET SKILL:
{skill_record}
\end{promptbox}

\paragraph{Construction-time joint quality review.}
This review differs from the independent post-construction quality audit in
Section~\ref{app:benchmark-validation-prompt}. It filters and revises candidate
records rather than estimating the pass rate of the finalized benchmark.

\begin{promptbox}[Construction-time joint quality assessment]
You are reviewing a candidate query--skill pair for an implicit
skill-retrieval benchmark. Evaluate both the request itself and its
relationship to the annotated target skill.

Keep the sample only when:
1. The request is natural, concrete, and sufficiently informative.
2. Completing it materially requires a specialized skill or workflow.
3. The required capability remains implicit rather than explicitly named.
4. The annotated target provides a stable and well-supported answer.

Reject simple writing or formatting tasks, broad advice, explicit tool or
skill mentions, weak target associations, target mismatches, and requests with
excessive ambiguity among plausible targets.

Use one of the following values:
- skill_necessity: high, medium, or low
- difficulty: hard, medium, or easy
- issue_type: good_implicit_reasoning, simple_task, vague_or_generic,
  explicit_match, target_mismatch, excessive_target_ambiguity, or other

Return only JSON:
{"keep":true,"skill_necessity":"high","difficulty":"hard",
"issue_type":"good_implicit_reasoning",
"reason":"one concise sentence","suggested_query":""}

CANDIDATE:
{query_target_and_metadata}
\end{promptbox}

\paragraph{Query-specific alternative-skill adjudication.}
A consistent capability-level substitutability criterion is used for both
construction-time ambiguity auditing and the alternative-skill sensitivity
analysis.

\begin{promptbox}[Query--pool alternative-skill adjudication]
You are reviewing whether a candidate skill is an acceptable alternative for
one specific benchmark query. Judge whether the candidate can satisfy this
request, not merely whether it shares a topic, terminology, or embedding
similarity with the annotated target.

Return ``positive'' only when the candidate can independently satisfy the
core request and can replace the annotated target without losing an essential
capability.

Return ``ignore'' when the candidate is related or potentially useful but
covers only a neighboring, broader, narrower, auxiliary, or partial
capability.

Return ``hard_negative'' when the similarity is superficial or the
candidate's capability boundary is incompatible with the request.

Do not promote a candidate solely because its cosine similarity is high.
Forks or copied implementations with the same capability are generally
positive, whereas same-domain skills with different capability boundaries
are generally ignore or hard_negative.

Return only JSON:
{"decision":"positive","reason":"one concise sentence"}

QUERY:
{query}

ANNOTATED TARGET SKILL:
{target_skill}

CANDIDATE SKILL:
{candidate_skill}
\end{promptbox}

\subsection{Training-Data Generation Prompts}
\label{app:training-data-prompts}

\paragraph{Stage-I query synthesis.}
\begin{promptbox}[Sonnet query synthesis]
Given the target skill record, write three distinct and realistic user
requests for which this skill is the correct target.

Requirements:
- Express a concrete functional need, including useful input context,
  constraints, or an expected deliverable.
- Write natural user requests rather than benchmark descriptions or skill
  summaries.
- Do not mention the skill ID, skill name, repository, or the phrase
  ``use this skill''.
- Do not directly restate the skill description.
- Do not copy a distinctive procedure from the skill document.
- Keep all three requests aligned with the same target capability.
- Make the three use cases meaningfully different from one another.

Return exactly one JSON object and no additional text:
{"queries":["first request","second request","third request"]}

TARGET SKILL:
{skill_name}
{skill_description}
{skill_document}
\end{promptbox}

\paragraph{Training-query quality review.}
\label{app:training-query-review-prompt}

\begin{promptbox}[Training-query quality review]
You are reviewing a synthetic user request for skill-retrieval training.
Evaluate whether the request is natural, informative, and correctly aligned
with the target skill.

Keep the request only when:
1. It describes a plausible and concrete user need.
2. The target skill provides the capability required to satisfy the request.
3. The request contains sufficient context, constraints, or a desired
   deliverable for retrieval.
4. It does not name, summarize, or directly copy the target skill.
5. It is meaningfully distinct from the other requests generated for the same
   target.

Reject requests that are malformed, unnatural, overly generic, redundant,
insufficiently informative, explicitly reveal the target, or require a
different capability.

Use one of the following issue types:
good, malformed, unnatural, generic, redundant, target_leakage,
target_mismatch, or insufficient_context.

Return only JSON:
{"keep":true,"issue_type":"good","reason":"one concise sentence"}

CANDIDATE REQUEST:
{query}

TARGET SKILL:
{skill_name}
{skill_description}
{skill_document}

OTHER REQUESTS FOR THE SAME TARGET:
{other_candidate_queries}
\end{promptbox}

\paragraph{Capability-level rationale annotation.}
\begin{promptbox}[Capability-level rationale annotation]
Given a user task and the known relevant skill capability, write one concise
query-analysis sentence for skill retrieval.

Requirements:
- State only the capabilities required by the user.
- Do not provide implementation steps, an extended explanation, or an answer
  to the task.
- Ground the capability description in the known relevant skill while adapting
  it to the specific user request.
- Do not mention skill names or skill IDs.
- Do not output JSON, markdown, bullets, numbered steps, or labels.
- Use exactly this sentence pattern:
  ``The user needs skills for <capability description>.''
- Keep the sentence under 32 words.

USER TASK:
{query}

KNOWN RELEVANT SKILL CAPABILITY:
{positive_skill_capability}
\end{promptbox}

\paragraph{Stage-II implicit-query rewriting.}
\begin{promptbox}[Stage-II implicit-query rewriting]
Rewrite the explicit skill-retrieval training query as a natural implicit user
request. Preserve the same underlying task and target skill while making the
request less dependent on explicit capability or tool terminology.

Requirements:
1. Preserve the task intent, available inputs, constraints, and expected
   deliverable.
2. Preserve the capability boundary of the original target skill.
3. Make the rewritten request sound like a plausible real-user request.
4. Prefer descriptions of goals and desired artifacts over explicit tool,
   algorithm, API, package, or capability names.
5. Retain enough task information for skill retrieval.
6. Do not mention the skill ID, skill name, or wording copied directly from
   the skill document.
7. Do not introduce requirements absent from the original request and target
   skill.
8. Return exactly one JSON object and no additional text.

Return only JSON:
{"implicit_query":"...","reason":"one concise sentence"}

INPUT:
{"original_query":"{query}",
"query_capability":"{capability_rationale}",
"positive_skill":"{skill_record}"}
\end{promptbox}

\subsection{Reranker Candidate-Relevance Audit Prompt}
\label{app:reranker-positive-audit-prompt}

\begin{promptbox}[Multi-skill candidate-relevance audit]
You are assessing whether one candidate skill is a valid supervised target for
a multi-skill agent request. Judge the concrete capability supplied by the
candidate and its contribution to this specific request. Do not infer
capabilities that are absent from the candidate document.

Assign exactly one label:

- strong_positive: The candidate directly provides a core capability required
  by the request or necessary for a central subtask. Without it, an essential
  task requirement remains uncovered.

- auxiliary_positive: The candidate provides a concrete supporting capability
  for a secondary subtask, but it is not a core routing target.

- borderline_positive: The candidate is plausibly useful, but the request or
  document provides insufficient evidence that its capability is required.

- weak_generic: The candidate offers only a broad or generic capability when
  the request requires a more specific tool, workflow, or domain procedure.

- wrong_positive: The candidate does not provide a capability required by the
  request, is mismatched, or addresses a different task.

Use auxiliary_positive only when the request contains a concrete secondary
subtask supported by the candidate. Do not assign a positive label solely
because of retrieval rank, embedding similarity, shared terminology, or broad
helpfulness. Base the decision on the documented capability and the explicit
requirements of the request.

Return only JSON:
{"label":"strong_positive","reason":"one concise sentence"}

USER REQUEST:
{query}

CANDIDATE SKILL NAME:
{skill_name}

CANDIDATE SKILL DESCRIPTION:
{skill_description}

CANDIDATE SKILL BODY:
{skill_body_excerpt}
\end{promptbox}

\SkillReasonEndSupplement


\end{document}